\documentclass[fleqn,10pt]{wlscirep}
\usepackage[utf8]{inputenc}
\usepackage[T1]{fontenc}
\usepackage{lineno}
\usepackage{makecell}
\usepackage{bm}
\title{ Neuromorphic Pseudo-Random Number Generators with a Low Power Hardware Implementation}

\author[1, 2, 3]{Jafar Shamsi}
\author[1]{Navid Akbari}
\author[5]{Sonia Sennik}
\author[1,4,*+]{Aaron Gruber}
\author[1,*+]{Wilten Nicola}
\affil[1]{Hotchkiss Brain Institute, University of Calgary, Calgary, Canada}
\affil[2]{Biomedical Engineering, University of Calgary, Calgary, Canada}
\affil[3]{Electrical and Software Engineering, University of Calgary, Calgary, Canada}
\affil[4]{ACHRI, University of Calgary, Calgary, Canada}
\affil[5]{Creative Destruction Lab}

\affil[*]{Corresponding Author: wilten.nicola@ucalgary.ca, aaron.gruber@ucalgary.ca}
\affil[+]{these authors contributed equally to this work}

\begin{abstract}
Pseudo-random number generation often requires trade-offs among quality, power consumption, and bandwidth to produce unpredictable sequences of numbers. The brain, on the other hand, efficiently generates unpredictable output complex network dynamics occurring in a high-dimensional state.  This state, which is hypothesized to be chaotic, relies on the balance between excitation and inhibition. Here, we investigated if computational models of these chaotic balanced states can be harnessed for Neuromorphic Pseudo-Random Number Generators (NPRNGs) in low power hardware.  We successfully constructed a balanced spiking neural network model consisting of leaky-integrate-and-fire neurons that could be readily implemented in low power FPGAs and used as a NPRNG.  The prototyped NPRNG consumed 3.24 mW during operation and produced pseudo-random numbers at 120kbps.  In both hardware and software instantiations, NPRNGs produce high-quality random numbers as validated by standard metrics for testing RNG quality.
\end{abstract}

\begin{document}

\flushbottom
\maketitle

\thispagestyle{empty}

\section*{Introduction}

Pseudo-random number generators (PRNGs) are important for modern computing. PRNGs are used to solve advanced computational problems \cite{mh}, or as inputs for generative artificial intelligence \cite{diffnets}.   Many PRNGs operate as a recurrence relation, where an initial state (the seed) is transformed by iteratively applied operations to produce random-looking deterministic streams of bits. Unfortunately, the complex operations that enable good pseudo-random number generation are often resource and power-intensive to implement in hardware like field-programmable gate arrays (FPGAs) or even in application-specific integrated circuit (ASIC) chips \cite{Silva2023,Cardoso2021,Kopparthi2022}.

In contrast, the brain produces unpredictability efficiently.   Experiments have shown that certain dynamical brain states are high-dimensional and likely chaotic \cite{destexhe1,london1}.  Further, the balance between excitation and inhibition measured in cortical brain networks is computationally modeled by strong, but balanced coupling between neurons which leads to chaotic dynamics \cite{ei1,ei2}.  This dynamical state, in which the neurons fire asynchronously, is readily reproduced with randomly generated synaptic weights in both recurrent artificial neural networks \cite{ei1,ei2,J1}, and recurrent spiking neural networks \cite{B1,B2,B3,F1,F2,F3,S1,N1,harish,Osc1}.  As such networks increase in size, it is mathematically provable that the constituent neurons activate as if they are increasingly independent and driven by a Gaussian process \cite{harish,N1,F1}.  Collectively, experimental and theoretical evidence suggest the possibility of efficient random number generation based on \textit{neuromorphic computing}, in which some aspects of brain computation, such as the firing of spikes by neurons, are emulated in a small hardware form factor.   Further, a native neuromorphic pseudo-random number generator would be useful for neuromorphic computing instantiations, as it could rely on existing neuromorphic primitives for implementation, rather than additional circuit elements.

In this work, we sought to harness brain-based computing principles to produce low-power neuromorphic pseudo-random number generators (NPRNGs).  First, we show that relatively small networks of integrate-and-fire neurons under a balanced connectivity regime \cite{ei1,ei2,N1,harish,S1,F1} produce sufficient entropy in a chaotic regime to act as good PRNGs, as validated with the National Institute of Standards and Technology (NIST) SP-800-22 testing package for PRNGs \cite{NIST}.  We found that not all configurations of balanced spiking networks were suitable for PRNGs. The connectivity parameter, "$g$", critically determined the quality of pseudo-random number generation.  If the connectivity strength was too large, the spiking neural networks transition to ``rate-chaos" regimes \cite{S1,harish} that led to poor pseudo-random bit streams.  If the connectivity strength was sufficiently large but less than a critical threshold, then spike-chaos emerged and small networks of neurons displayed Poisson-like firing statistics. These spikes produced high quality pseudo-random number bit streams that matched or exceeded other established PRNGs.  Due to the binary weights, each NPRNG was unique and occupied a large parameter space of $O(2^{N^2})$.  This also allows an NPRNG to be reconfigured by disabling neurons with inhibitory bias currents.  We next implemented the best NPRNGs in hardware with low-power FPGAs.  These fabricated NPRNG devices generated random numbers at 120kbps while using 5.45 mW of power, considering all circuit components during operation. The power consumption reported by Lattice Radiant Software was 3.24 mW, which only includes the FPGA device without peripheral components.  To the best of our knowledge, this work represents a novel application of chaotic balanced SNNs as a basis for pseudo-random number generation.

\section* {Results}

\subsection*{The excitatory/inhibitory balanced state in quantized spiking neural networks.}

To investigate the capacity of chaotic spiking neural networks (SNNs) to serve as pseudo-random number generators, we considered balanced networks of leaky-integrate-and-fire (LIF) neurons:
\begin{eqnarray}
\tau_m\dot{v}_i &=& - v_i +\sum_{j=1}^N w_{ij}r_j(t) + I, \quad i=1,2,\ldots N  \label{snn1},\\
\dot{r}_j(t) &=&  -\frac{r_j}{\tau_d} + h_j\\
 \dot{h}_j(t)&=& -\frac{h_j}{\tau_r} + \frac{1}{\tau_d\tau_r}\sum_{t<t_{jk}}\delta(t-t_{jk})
\end{eqnarray}
where $v_i(t)$ corresponds to the voltage of the $i$th neuron while $r_i(t)$ and $h_i(t)$ implement the synaptic variables that allow neurons to communicate with one another.  A neuron is said to fire a spike when it reaches a threshold $v_i(t^-) =v_{th}$ and is reset at $v_i(t^+)=v_{reset}$.  The spikes are filtered by a double-exponential post-synaptic filter ($r_j(t)$) for each neuron and then multiplied by the weight matrix $\bm w$ to form a network.  The LIF model was chosen because it is simple to apply in hardware and is considered a computational primitive in neuromorphic systems \cite{Frenkel2023,Pedersen2024}.  The background current $I$ drives an isolated neuron, $v_i(t)$, to fire spikes (Materials and Methods) and is set at the threshold for firing $I = v_{th}$ in the balanced configuration.  Neurons can also be disabled by setting $I\ll v_{th}$.

Depending on the coupling weights, $\bm w$, the SNN in equation (\ref{snn1}) can display complex dynamics \cite{harish,S1,F1}.  These complex dynamics will be harnessed for pseudo-random number generation by applying a transform $T(s)$ to the resultant spike-train to generate pseudo-random bits. In the balanced state, the excitatory weights ($w_{ij}>0$) are counteracted by the inhibitory weights ($w_{ij}<0$), by imposing the following conditions on the first two moments of $w_{ij}$:
\begin{eqnarray}
E(\omega_{ij})  = 0,\quad E(\omega_{ij}^2) \propto \frac{1}{N} \label{bnn1}.
\end{eqnarray}
where $E$ is the expectation operator.  More simply phrased, the mean of the weights is set to 0, while the variance of the weights scales like the inverse of the system size. The parameter regime in (\ref{bnn1}) is a well-studied coupling regime for recurrent neural networks.  This coupling regime produces chaotic dynamics wherein the neurons asymptotically ($N\rightarrow\infty$)  behave as if they were uncoupled, and independently driven by a filtered noise process.  This noise process can be self-consistently derived and depends on the characteristics of the coupling and neuronal dynamics \cite{F1,F2,F3,S1,S2,N1,Osc1,B1,B2,B3}.  

Here, we investigate the potential for the irregular spiking dynamics of balanced networks (\ref{snn1})-(\ref{bnn1}) as a basis for neuromorphic pseudo-random number generators (NPRNGs).    The networks considered will be modified to be implementable in low-power hardware (FPGAs).  

\begin{figure}[htp!]
    \includegraphics[scale=0.4]{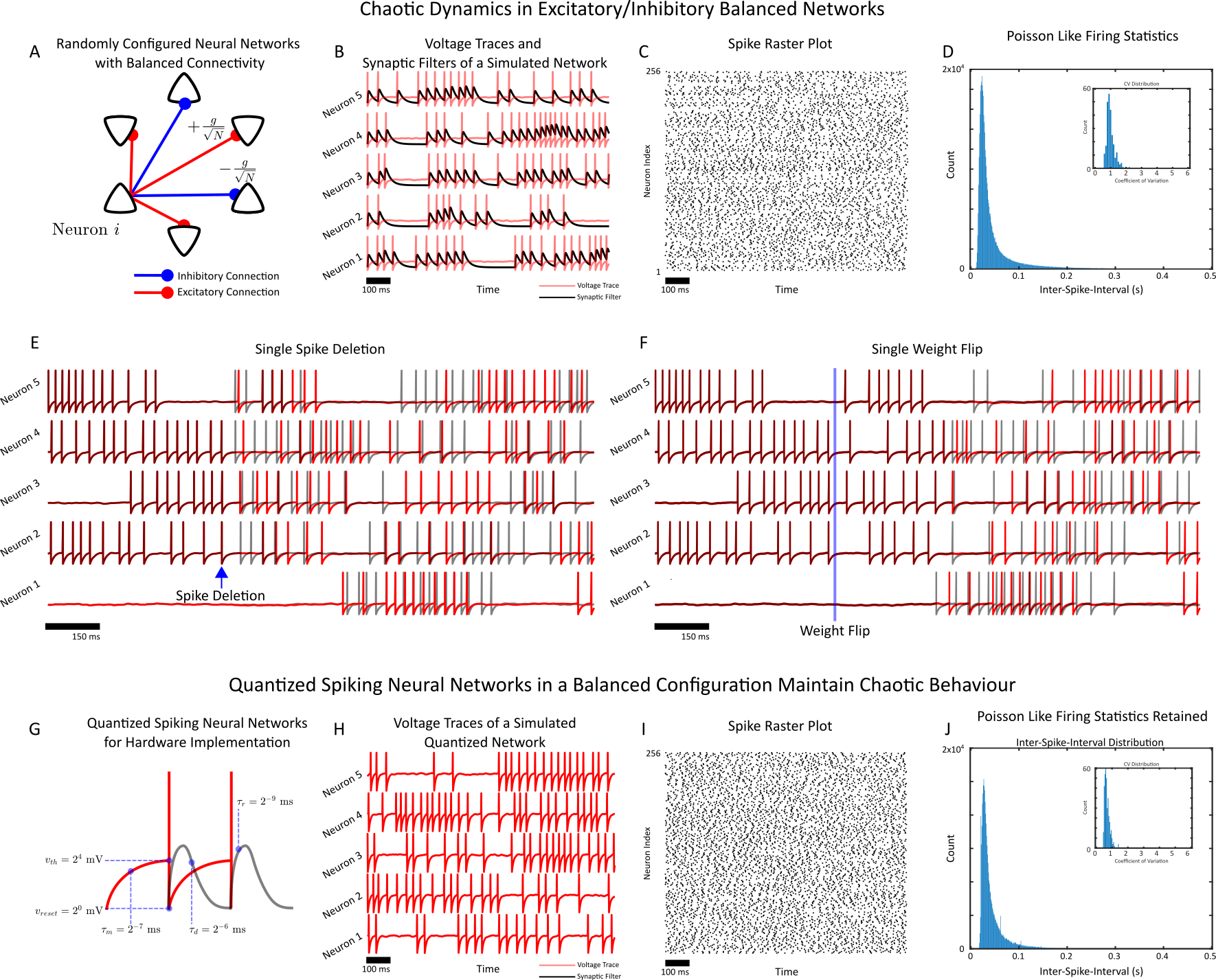}
    \caption{Quantizing-Balanced Spiking Neural Networks for Pseudo-Random Number Generation.  \textbf{(A)} Schematic of a spiking neural network with random, but balanced connectivity.  Neuron $i$ forms excitatory ($+\frac{g}{\sqrt{N}}$) and inhibitory synaptic  ($-\frac{g}{\sqrt{N}}$) connections.  
\textbf{(B)} The voltage traces (red) and postsynaptic filters (black) for 5 neurons in a simulated balanced network connected as in (A) for 1 second of simulation time. 
\textbf{(C)}  A spike raster plot for 1 second of simulation for all $N=256$ neurons. 
\textbf{(D)} The inter-spike-interval (ISI) histogram measured for all neurons in a 100 second simulation of a network with $N=256$ neurons (from (B)-(C)).  The lack of density at 0 is caused by the finite time required for a neuron to generate a spike by integrating a current (the relative refractory period).  The distribution of the coefficient of variation (CV) for each of the $N=256$ is shown (inset). A CV of 1 indicates spiking behaviour similar to that of a Poisson process. 
\textbf{(E)} The irregular dynamics exhibited by balanced networks are chaotic: if a single spike fails to produce a post-synaptic effect, the subsequent constellation of spikes produced by the network is altered.  The non-deleted voltage trace is shown in red, while a spike deletion is performed at 500 ms, for a single neuron (blue).  
\textbf{(F)} The irregular dynamics exhibited by balanced networks are sensitive to single weight flips.  For a pair of parameter-matched, parallel simulations, a single weight ($w_{12}$ ) was flipped in sign (excitatory to inhibitory).   The voltage trace for the original weight matrix is shown in red, whereas the grey traces show activity in the network before and after the weight flip (vertical line).   
\textbf{(G)}  In a quantized version of the spiking neural network, all network parameters are set to powers of 2 (binary) to enable a simplified implementation of all network operations in hardware. 
\textbf{(H)} A 1-second simulation of a quantized, balanced neural network, with the voltage traces for 5 neurons was plotted.  
\textbf{(I)} A 1-second raster plot of a quantized, balanced neural network, for all $N=256$ neurons.  
\textbf{(J)} The ISI distribution, and CV distribution (inset) for all of the spikes in a quantized, balanced neural network. 
} 
\label{figure1}
\end{figure}

The first such modification is weight binarization:
\begin{eqnarray}
\omega_{ij} = \pm \frac{g}{\sqrt{N}}
\end{eqnarray}
where $g$ is the overall coupling strength of the connections (Figure \ref{figure1}A).  This modification reduces the storage required for on-chip implementation and was previously considered in other studies \cite{N1}.  To investigate if binarizing the weights impacts the balanced state, we ran a series of simulations with a network consisting of $N=256$ neurons.  We confirmed that the irregular voltage and spiking dynamics (Figure \ref{figure1}A-B) are still present and the neurons fire spikes with a distribution of inter-spike-intervals (ISIs) that is similar to a Poisson process with a refractory period (Figure \ref{figure1}D).  The ISIs have a mean coefficient of variation of $\mu_{cv} =0.99$ ($N=256$, $\sigma_{cv} = 0.25$, Materials and Methods), indicating Poisson-like spiking.  Further, for larger values of $g$, an SNN coupled with binarized weights still displays a transition to ``rate-chaos" \cite{S1,harish} (Supplementary Figure S1).

Next, we tested if the network's state trajectory was altered by the deletion of a single spike, which is a necessary feature of chaotic SNNs \cite{S1,F3,B2}.  Indeed, deleting a single spike (Figure \ref{figure1}E) led to a completely different constellation of spikes, as compared to a parameter-matched and initial-state-matched simulation without deletion.  The SNN with binarized weights also displays a sensitivity to the underlying parameters, like any chaotic system, in which changing a single connection leads to a different irregular trajectory (Figure \ref{figure1}F). 

Next, we considered quantizing all network parameters to facilitate efficient hardware implementation.  Quantization entails finding parameters that are of the form $\pm 2^k$ where $k$ is an integer (Figure \ref{figure1}G) for all neuronal and network parameters.  This includes the binary coupling strength $g/\sqrt{N}$.  Quantization is critical for an efficient hardware implementation of the SNN because it converts all operations to simple movements of bits.  The quantized network still displays irregular spiking dynamics (Figure \ref{figure1}H-I) in which neurons display similar Poisson-like spiking as in the non-quantized network (Figure \ref{figure1}J), although with a smaller coefficient of variation for the parameters considered ($\mu_{CV}=0.68$, $\sigma_{CV} = 0.16$, $N=256$). The quantized network also displays a transition to ``rate chaos" as the coupling strength is varied (Supplementary Figure 2).  In summary, the key modifications required to implement a balanced network in neuromorphic hardware (quantization and weight binarization) do not impact the overall dynamical features of the chaotic state.   We remark that in a system that is discretized, as the SNN considered here, the simulated dynamics are only an approximation to the true chaotic system analyzed mathematically by others (e.g. \cite{S1,harish}).  The system will eventually cycle, and thus we include a derivation of the maximum and expected cycle time in Supplementary Section S4. 

\clearpage

\clearpage

\subsection*{Creating and Testing Random Bit Streams with Balanced SNNs}

We next considered mechanisms for translating the irregular firing of neurons in the SNN into a bit string for use as a string of pseudo-random numbers.  An ideal pseudo-random number generator produces its $n$th output bit ($b_n$) as if it was independently drawn from an unbiased Bernoulli random variable:
\begin{eqnarray}
b_n = \begin{cases} 0 & \text{with probability } \frac{1}{2} \\ 1 & \text{with probability }  \frac{1}{2} \end{cases} 
\end{eqnarray}
with $b_j$ and $b_k$ being statistically independent for any $j\neq k$.   

\begin{figure}[htp!]
\centering
    \includegraphics[scale=0.55]{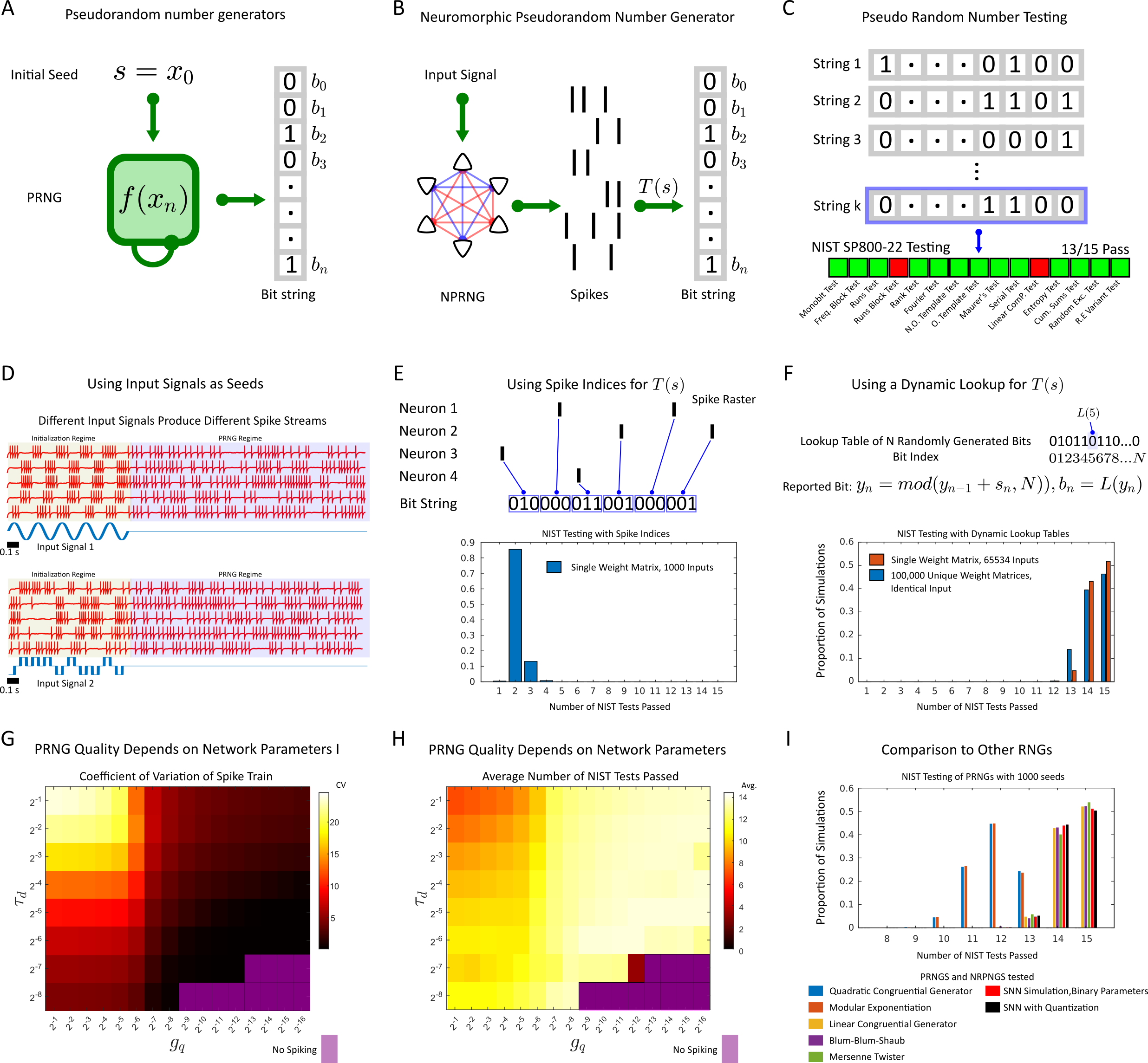}
    \caption{Using Balanced Spiking Neural Networks to Generate Random Bit Strings.  \textbf{(A)} Many conventional pseudo-random number generators (PRNGs) use an input seed $s$ and a recursive/iterative operation $f(x)$ to generate a bit string.  
\textbf{(B)} A neuromorphic pseudorandom number generator (NPRNG) operates differently from a conventional PRNG.  Due to hardware limitations, the neural network requires an input signal to initialize the system onto a chaotic trajectory, and generate spikes/spike times pseudorandomly.  The NPRNG produces a spike raster plot, which is then transformed into a bit stream with some mapping function, $T(\bm s)$.
\textbf{(C)} The NIST-SP800-22 package of statistical tests is used to test the null hypothesis that the bit string is generated from a series of independent Bernoulli random variables with $b_k = 1$ with probability $p=1/2$, and 0 otherwise (see Methods for a detailed discussion).   
\textbf{(D)}   (Top) A 5 Hz sinusoid is used as the input signal for a quantized network of $N=256$ neurons.  The signal is shut off after 1 second.  (Bottom) A randomly generated sequence of square wave pulses is used as a second input signal for the same network as in (A). 
\textbf{(E)} NIST testing of a a naive transform, $
\bm b =T(\bm s)$ for converting the spike raster plot into a bit stream. For a network of $N=2^k$ neurons in base $k$, each spike index is represented in $k$-bit binary.   NIST testing reveals this as a poor way of generating pseudo-random bits.
\textbf{(F)} NIST testing of NPRNG bit streams with dynamic look-up tables for a single weight matrix with different inputs (orange) and different weight matrices (blue).  
\textbf{(G)} The coefficient of variation of the inter-spike-interval distribution, as a function of the coupling $g_q$ (x-axis) and the decay time, $\tau_d$ (y-axis) of the synapses. The bright region in the top-corner corresponds to the ``rate-chaos" regime for the SNN.  
\textbf{(H)} The average number of NIST tests passed as a function of the coupling $g_q$ (x-axis) and the decay time, $\tau_d$ (y-axis) of the synapses.  
\textbf{(I)} Comparing quantized and non-quantized SNN bitstreams to bitstreams generated by other PRNGs. 
} \label{figure2}
\end{figure}

Many conventional pseudo-random number generators (PRNG) operate by using an initial input, or seed, $s=x_0$ and some recursive operation, $x_{n+1}=f(x_n)$ to produce pseudo-random sequences of bits (Figure \ref{figure2}A).  For example, the Blum-Blum-Shub (BBS) PRNG \cite{bbs} uses the nonlinear recurrence relationship 
\begin{eqnarray}
x_{n+1} = x_{n}^2\mod M
\end{eqnarray}
where $M = pq$ is the product of two primes and $p$ and $q$ are both Blum primes.  When converted into bits, the least significant bits of $x_{n+1}$ produces a pseudo-random bit string. 

A neuromorphic pseudo-random number generator NPRNG, has two key differences with conventional PRNGs (Figure \ref{figure2}B).  First, an NPRNG requires an initial state for more variables, in particular each voltage, $v_i(t)$ and synaptic filters $r_i(t)$.  This produces complexity for hardware implementation due to the resource (area) limitations of low-power chips.   However, a 1D time series can also be used as an input signal to initialize an NPRNG onto a chaotic spiking trajectory (Figure \ref{figure2}B) and serve as a ``seed" when all of the initial conditions ($v_i(0)$ and $r_i(0)$) are hard-coded and constant.  Alternatively, an $N$-bit binary seed can initialize the network where the $j$th neuron's initial condition is set to super-threshold (spiking) if the $j$th bit of an input seed is 1.  If the $j$th bit of the input seed is 0, the initial condition for the neuron is $v_i(t)=v_{reset}$.  We use this initialization later when we consider hardware-based NPRNGs.

The second difference between an NPRNG and a conventional PRNG is the need to transform the spike-time sequence of all neurons, which we label with the $n\times 2$ matrix $\bm s$, where $s_{n1} = j_n$, the index of the neuron that fired the $j$th spike, and $s_{n2} =t_n$, the time at which the $j$th spike is fired. We transformed the time series of spike times into a bit string by a process described below.

We utilized a series of 15 statistical tests developed by the National Institute of Standards and Technology (NIST, NIST-SP-800-22,\cite{NIST}, Materials and Methods) to determine the quality of pseudo-random number bit strings generated by NPRNGs (Figure \ref{figure2}C).  These tests determine if a bit string is indistinguishable from a random one via various measures (e.g. repetition of sequences, frequency of bits, etc.).  The average number of tests passed was considered for 100 randomly generated bit strings, each with 1,000,000 bits. 

Using the number of NIST tests passed as a metric for PRNG quality, we first investigated the use of input signals, rather than predefined initial states as seeds.  These input signals allow a streaming input to initialize the network onto a chaotic trajectory (Figure \ref{figure2}D). The input signal is applied to each neuron with a binary weight matrix:
\begin{eqnarray}
\tau_m\dot{v}_i &=& - v_i +\sum_{j=1}^N w_{ij}r_j(t) + I + w^{in}_i c(t), \quad i=1,2,\ldots N  
\end{eqnarray}
where $c(t)$ is the input signal, and $w^{in}_i = \pm 1$ is a randomly generated weight.  With this implementation, only $O(N)$ bits of storage (for $\bm w^{in}$) are required for the hardware implementation. Two separate input signals $c_1(t)$ and $c_2(t)$ can cause a network with the same initial conditions ($v_i(0) = v_{reset}, r_i(0)=0$, $i=1,2,
\ldots N$) to produce two different chaotic trajectories, and therefore, bit strings.  Further, inputs can also be used to synchronize two NPRNGs that are on initially different chaotic trajectories, provided that $c(t)$ is of sufficiently long duration (Supplementary Figure S3). This is partially a consequence of the stable but irregular nature of the chaotic trajectories (stable-chaos) displayed by balanced SNNs \cite{F1,F2,F3}.  

With a mechanism for seeding an NPRNG determined, we next considered how to generate bits with the spike raster through some transform, $T(\bm s)$.  One naive transform would be to take the $n$th spike index for a network of $2^k$ neurons and convert it into a binary representation with $k$-bits, and group the resulting $nk$ bits together into a bitstream.  This turns out to produce bitstreams that are easily detectable by NIST as being different from the independent Bernoulli random bitstream hypothesis. In particular, this occurs because a neuron in a balanced network has a kind of relative refractory period that follows each spike in which the neuron is unlikely to spike again (Supplementary Figure S4).  This refractory period is caused by the integration time constant of the neuron $\tau_m$ controlling how quickly a neuron can reach threshold after being reset by integrating its inputs.  This produces bit streams that appear as non-random, as the bits generated by a neuron's index inherit the neuron's refractory period.  

To produce higher-quality PRNGs, we implemented a transform $T(\bm s)$ that would take into account the relative refractory period of firing.  First, a counter variable $y_n$ is computed for the $n$th spike:
\begin{eqnarray}
y_n = \text{mod}(y_{n-1}+s_n,N)
\end{eqnarray}
This variable accumulates the previous history of spiking by the entire network. Next, we define a static lookup table $L(i)$, $i=1,2,\ldots N$ which corresponds to $N$ uniformly distributed bits.  The $n$th reported bit is then given by 
\begin{eqnarray}
b_ n = L(y_n)
\end{eqnarray}
The bits in the lookup table are static, and are only ever generated once.  The lookup table $L$, recurrent weight matrix $\bm w$, and input weight matrix $w_{ij}$ collectively define an NPRNG.  

We tested this simple-to-implement and memory-efficient transform $T(\bm s)$ with a single randomly generated weight matrix, with 65,536 separate inputs (Materials and Methods).  Each input generated a bitstream consisting of $10^8$ bits.  These $10^8$ bits were divided into 100 individual streams of $10^6$ bits each, and were tested with NIST (Figure \ref{figure2}F).  Approximately 50\% of all initial inputs led to bitstreams that passed every NIST test, with another ~40\% passing 14/15 NIST tests, and only a small number of bitstreams passing 11-13 of 15 tests. 

Next, we considered the impact of the individual weight matrix.  We generated $10^4$ separate weight matrices and produced a bitstream of $10^8$ bits for each weight (Figure \ref{figure2}F).  We found that the population of weights on average produced slightly poorer quality bitstreams than the single randomly generated weight matrix we considered, indicating some impact on the specific weight matrix generated for an NPRNG.  However, the overall distribution of passed NIST tests is qualitatively similar across weights, with no generated weight matrix having failed less than 10 tests or fewer.  Additionally, approximately half of all weights passed all 15 NIST tests.  We remark that it is possible to generate a non-random, but still technically ``balanced" weight matrix that will generate poor-quality random bitstreams, while still satisfying the conditions in (\ref{bnn1}) (Supplementary Figure S6).  The key assumption in Supplementary Figure S6 that is violated is the statistical independence of the individual weights $w_{ij}$.  Additionally, we also remark that networks that are too small ($N<64$) may also have stable non-chaotic attractors with sizable basins of attraction.  This was confirmed by numerically estimating the maximum Lyapunov Exponent of SNNs for increasingly larger $N$ along with NIST Testing of simulated networks of size $N=128,256,512$ (Supplementary Figure S5, Supplementary Figure S8).  

While balanced SNNs are chaotic, the chaos can be of two different forms: rate or spike chaos \cite{S1,harish}.  In rate chaos, the neuronal firing rates have long time-scale autocorrelation functions where neurons tend to fire bursts of spikes with time-varying rates, rather than the isolated, Poisson-like spiking in spike-chaos \cite{S1,harish}.  We sought to determine how the statistics of spiking in these different chaotic states impacted the quality of pseudo-random bit generation.  We simulated networks of varying connection strengths, $g_q = 2^{-j}$ for $j=1,2,\ldots 16$ and varying synaptic time constants $\tau_D = 2^{-i}$, for $i=1,2,\ldots 8$, and computed the mean coefficient of variation for the spike rasters generated (Materials and Methods, Figure \ref{figure1}G).  Note that the $\sqrt{N}^{-1}$ is absorbed into $g_q$.  Over this parameter range, we found a coexistence between non-firing (small $g_q, \tau_D$), rate chaos (larger $g_q$) and Poisson-like spiking (smaller $g_q$) where the CVs are small $\mu_{cv}<5$.   Next, for the same parameter regimes, we computed the average number of NIST tests passed (Figure \ref{figure2}H).  In the rate-chaos regime, where the coefficient of variation is high, poorer quality bitstreams are produced (Figure \ref{figure2}H).  The average number of NIST tests passed was also strongly negatively correlated $(\rho = -0.8147$, $p\ll 10^{-4}$), with the average coefficient of variation $\mu_{cv}$ (Supplementary Figure S5). Conceptually, this significant difference in performance is rooted in the temporal correlations of the two chaotic regimes. In the rate-chaos regime, caused by excessive coupling parameters, the network is dominated by slow and long fluctuations where neurons fire in extended bursts. This behavior generates strong temporal autocorrelations \cite{harish,S1}. When these burst-dominated spike patterns are processed by our dynamic lookup table, the resulting bitstreams inherit these temporal dependencies, creating detectable non-random patterns that fail statistical testing. Conversely, the spike-chaos regime operates as an asynchronous firing state where fluctuating excitatory and inhibitory inputs balance dynamically. This balancing forces individual neurons to fire isolated spikes with Poisson-like inter-spike-interval statistics \cite{F1}. Because the autocorrelation function remains flat and cross-correlations are weak in this regime, successive spiking events are statistically independent in the large $N$-limit, enabling the dynamic lookup table to reliably generate unbiased, seemingly independent Bernoulli bitstreams. Collectively, these results imply that the Poisson-like spiking regime is superior to the rate-chaos regime for generating random bitstreams with the dynamic lookup table $T(\bm s)$.  However, we remark that the rate-chaos regime may still be used in the NPRNG in future applications, potentially by using other transforms $T(\bm s)$.  

One potential issue that arises with simulated chaotic systems is that when discretized in both space and time, cycling must occur eventually.  True chaos is defined over a continuum in space in discrete time dynamical systems and a continuum in time and space in a continuous time dynamical system.  To check for cycles, we simulated a quantized NPRNG which generated 1.1 TB of random bits, each bit corresponding to a single spike.  We used the last 1000 bits and searched over the random bit string and found no repetitions.  Further, a heuristic estimate using the birthday bounds demonstrates cycling is not likely to occur for any reasonable amount of time with 32 bits of precision used to approximate the variables of an NPRNG with 256 neurons (expected cycle time $\sim2^{12000}$, Supplementary Section S4).  This is in line with recent numerical results showing that simulated networks of chaotic systems have cycle times that grow geometrically with the system size \cite{cycle1}.  

Finally, to place these results into context, we compared how both quantized and non-quantized NPRNGs to other PRNGs.  We generated $10^4$ different 100 bitstreams, each containing $10^6$ bits, with different seeds for each stream. The streams were produced with: the quadratic congruential PRNG \cite{QC
, randu}, the modular exponentiation PRNG \cite{MEXP}, the linear congruential PRNG \cite{LQ1}, the Blum-Blum-Shub PRNG \cite{bbs}, and the Mersenne Twister \cite{MT}. The quadratic congruential generator and modular exponentiation performed the worst, with no bit streams generated passing all 15 NIST tests, while the LCG, BBS, MT, and the two NPRNGs all performed comparably, with $\approx 50\%$ of all bitstreams generated passing all 15 NIST tests, an additional $\approx 40\%$ passing 14 NIST tests, and the remaining $10\%$ of bit streams passing 10-13 tests.  These results show that balanced spiking networks serve as good PRNGs provided that the coupling strength between neurons is not too strong, and the neurons operate in the Poisson-like spiking regime.    

Next, we considered what unique features a spiking neural network PRNG would have.  First, we remark that the chaotic balanced state used as an NPRNG implementation is also simultaneously a reservoir in reservoir computing \cite{maass1,res1,res2}.  The complex nonlinear dynamics can be stabilized with a low rank perturbation to the existing weight matrix \cite{nicola1,nicola2,force1}.   However, NPRNGs are also reconfigurable as pseudo-random number generators.  In particular, we found that an NPRNG could output different bitstreams as disabling a few neurons with an inhibitory bias current changes the chaotic trajectory that the spiking neural network displays, but does not appreciably alter the statistics of spiking (Supplementary Figure S9).  Indeed, disabling neurons still results in passed NIST testing throughout identical areas of parameter space for a network of $N=256$ neurons with random subsets of 3 neurons turned off. Indeed, so long as not too many neurons are disabled, chaotic balanced spiking neural networks can still serve as NPRNGs so long as >128 neurons are still active (Supplementary Figure S8).  

To complement the standard NIST package and rigorously evaluate potential short-range dependencies across high-volume bitstreams, we extended our benchmarking to include two additional independent statistical suites: Dieharder and TestU01 (SmallCrush battery). Both batteries were evaluated across 100 distinct binarized weight matrices instantiated at our primary operating parameter set ($d_t=2^{-6}$, $g_s2^{-14}$). Under the TestU01 SmallCrush battery, all 100 structural weight configurations passed across every test depth; across 6,000 evaluated test statistics, the rate of flagged metrics remained strictly at the false-positive floor expected by statistical chance, with no catastrophic $p$-value failures or matrix-specific defects. Conversely, initial Dieharder evaluation of the unconditioned, raw model output revealed configuration-dependent failures in a subset of weight matrices, with defects concentrated exclusively in the short-range serial-correlation test family (e.g. \texttt{sts\_{serial}}, \texttt{rgb\_{lagged\_sum}}, \texttt{sts\_{runs}}). Short-range serial correlation can appear in raw outputs from physical and chaotic neural entropy sources, motivating lightweight digital conditioning such as XOR-based post-processing \cite{sato2021consideration, Li2023AppliedPhysics, Naveen2026, Yu2021Design, GarzonGonzalez2025, Silva2023}.
To condition the output, we applied nonoverlapping adjacent-bit XOR whitening implemented as recursive pairwise decimation $b_i \leftarrow b_{2i} \oplus b_{2i+1}$, a strategy equivalent to computing the parity of $2^k$-bit blocks as used in recent TRNG post-processing \cite{jia2026widening,vodenicarevic2017low}, and evaluated one, two, and three decimation passes. One pass was insufficient to remove short-range structure, whereas two passes eliminated the systematic Dieharder correlation failures. Three passes further stabilized the bitstreams, yielding only 4 isolated single-test hits across 11,400 tests, consistent with the expected false-positive rate for ideal random sources.

\clearpage 
\clearpage

\subsection*{NPRNGs in hardware run on approximately 5 mW of power.}

We prototyped a quantized/binarized NPRNG model on an FPGA-based device as a proof of concept. We selected a low power FPGA, iCE40UP5K, from Lattice Semiconductor, which dictated the area resource constraints (the available FPGA resources). 

\begin{figure}[htp!]
\centering
    \includegraphics[scale=0.45]{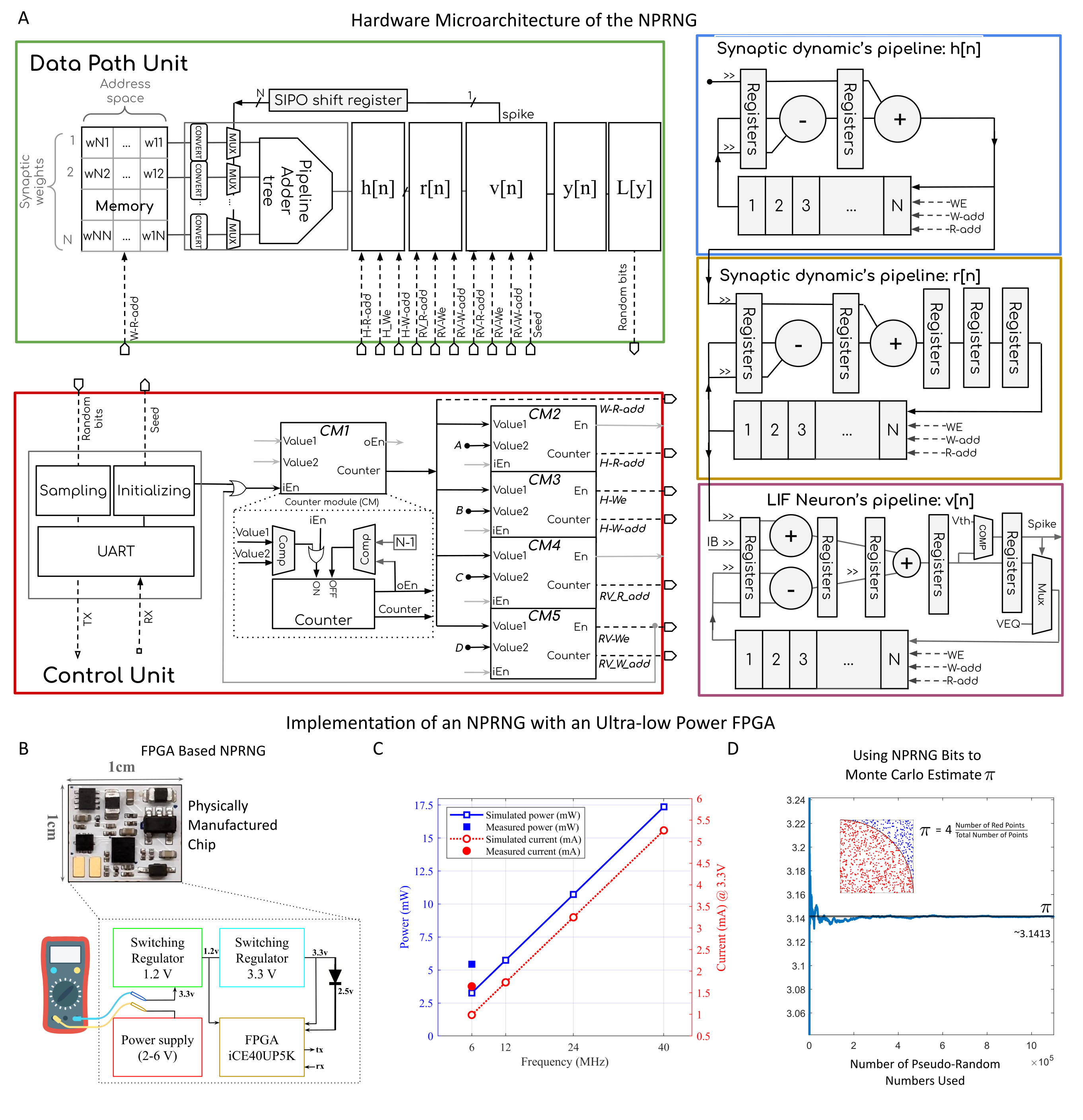}
    \caption{NPRNGs implemented in hardware with low Power Field Programmable Gate Arrays (FPGAs).  
\textbf{(A)}  Microarchitecture of the NPRNG.   (Left) Hardware microarchitecture of the NPRNG. The main microarchitecture (Left)  comprises a data path and a control unit. The data path unit includes a memory, multiply-accumulate (MAC) block, synaptic and neuron dynamics, function block $F[n]$, and the look-up table $L[y]$. The synaptic and neuron dynamics are implemented using pipeline design. Two, five, and five pipeline stages are used for $h[n]$, $r[n]$, and $v[n]$, respectively. Each block includes a computational block and a RAM. The synapse and neuron blocks perform the computations and the results are stored in the RAM for the next time step.
\textbf{(B)} Photograph of the 1cm $\times$ 1cm printed circuit board (PCB) with FPGA, two switching regulators, and one diode. 
\textbf{(C)}  Current and power consumption from simulations and tests. The higher power consumption on the fabricated device is due to the extra components and regulator inefficiencies.
\textbf{(D)} A Monte-Carlo estimate of $\pi$ using the hardware-based NPRNG.  A sequence of 2.2 million random numbers from the uniform distribution on $[0,1]$ were generated using the NPRNG.  These created 1.1 million ordered pairs, uniformly distributed on $[0,1]^2$ (the unit square).  The proportion of points that lie below the red line is approximately $\pi/4$.  The final Monte-Carlo estimate of $\pi$ using the NPRNG was 3.1413.   } \label{figure3}
\end{figure}

In the microarchitecture design, multiple techniques were applied to meet the area (FPGA resources) and performance (throughput) requirements (Figure \ref{figure3}A). To reduce area, the microarchitecture used shared hardware blocks for the synaptic dynamics ($h[n]$ and $r[n]$) and LIF neurons ($v[n]$) to reduce resource usage. Moreover, since spikes were used in the data representation, the synaptic multiplications were converted into additions, resulting in further reduction of resource usage. In addition, multiplications in the blocks $h[n]$, $r[n]$, and $v[n]$ were realized using shift and add operators resulting in a multiplier-less implementation. 

To improve the throughput, spatial parallelism (the Tree-Adder structure) was used for multiply-accumulate (MAC) block by performing the weight multiplications of each neuron in parallel (Supplementary Section S3).  The temporal parallelism (pipeline design) allowed an overall improvement of the throughput so that the output of one neuron is calculated on each clock cycle. The fully pipeline design also simplified the control unit, which was designed based on a modular counter-based block. This allows scalability of the design without making major changes in the microarchitecture but by setting parameters. The scalable design approach allowed us to configure the network size to fit the FPGA. A network of $N=256$ neurons was implemented utilizing 76 percent of the SLICEs (look-up tables and flip-flops) on the FPGA iCE40UP5K. The maximum clock frequency of around 41 MHz was estimated for this FPGA by the software of Synopsys Synplify Pro - Lattice Edition. To minimize the power consumption, the minimum frequency of the built-in oscillator, 6 MHz, was used. According to the power calculator of Lattice Radiant Software, the power consumption of the FPGA was as low as 3.24 mW at 6MHz (Figure \ref{figure3}D). The throughput of the NPRNG was estimated through simulation of the design using the software of Siemens ModelSim Lattice FPGA Edition. The NPRNG was simulated for 2000 time steps and the spike rate was calculated as $\approx$1/50 spikes per clock by dividing the number of generated spikes by the total clock cycles. Since a random bit is generated per spike, the throughput of the NPRNG is the spike rate multiplied by the clock frequency, 6MHz, resulting in 120kbps. The generated random bits were sampled every 5 bits before sending to the external device. To communicate with the external devices, the universal asynchronous receiver-transmitter protocol (UART) interface was used.

To measure the power consumption during the operation, the current of the board was monitored as 1.65 mA at 3.3 v resulting in 5.45 mW of the total power consumption (Figure \ref{figure3}C).  Finally, the random bits generated by the hardware-based NPRNG were also used to approximate $\pi$ with a Monte-Carlo method (Figure \ref{figure3}D), with the resulting approximation being $3.1413$ using 2.2 million randomly generated numbers from a uniform distribution on $[0,1]$.  

A comparison of the FPGA implementation of NPRNG and other PRNGs is provided in Table \ref{table_comparison}. Alongside  LCG-based and BBS-based PRNGs, implementations derived from chaotic maps and chaotic neural networks (cNNs) are incorporated for comparative analysis. Although this comparison table is not exhaustive, it does include selected state-of-the-art implementations representative of various PRNG categories. As a reconfigurable N-dimensional dynamical system, NPRNG uses more resources in terms of LUTs and FFs compared to LCG, BBS, and chaotic map-based PRNGs. Moreover, 25 BRAMs (125 Kbit) were used for storing the synaptic weights, the look-up table, and the internal states of the synapses and neurons. As a core methodology in translation of models to hardware, our approach eliminates dedicated multipliers entirely by mapping the SNN dynamics to distributed shift-and-add operators.In terms of power utilization, the absolute power consumption in this prototype can be as low as 3.2 mW.  This is within the range that can be passively powered by wireless energy sources \cite{Lazaro2020} (e.g. NFC), and may be useful for certain applications like generating the keys used for cryptographic algorithms. Although this FPGA version is meant for prototyping, the power usage and area of the ASIC design can be further optimized.

Our NPRNG architecture establishes a novelty boundary relative to these traditional chaotic map and Hopfield network paradigms. Algorithmically, conventional cNN designs rely on low-dimensional continuous trajectories where bits are extracted via word-level Least Significant Bit (LSB) truncation. Conversely, our architecture utilizes high-dimensional, event-driven spike-chaos, where entropy is natively derived from irregular inter-spike intervals through a dynamic lookup table that tracks the spatial-temporal history of the network. Microarchitecturally, prior hardware realizations require complex, power-intensive multiplier blocks or CORDIC (COordinate Rotation DIgital Computer)  units to compute continuous state activation variables. By implementing strict weight binarization and parameter quantization, our algorithm-hardware co-design converts the continuous mathematical system entirely into distributed shift-and-add operators with a spatial Tree-Adder unit. This allows a complex 256-dimensional neural state-space to operate inside a minimal digital footprint without utilizing a single dedicated hardware DSP multiplier. 

We distinguish this class of algorithmic PRNGs from device-level hardware primitives such as Intel’s DRAND or AMD’s cryptographic hardware keys. While the latter leverage localized analog physical non-determinism at the silicon device level, the listed PRNGs are algorithmic solutions, which can be implemented entirely within digital fabric as well as software. Under standardized NIST SP-800-22 metrics, both paradigms successfully satisfy the null hypothesis of indistinguishable randomness; however, our approach features a reconfigurable high-dimensional state-space and its seamless integration into neuromorphic processors. Furthermore, we differentiate our algorithmic solution from alternative high-speed circuit-level chaos primitives, such as Asymmetric Boolean Chaos Oscillators \cite{munoz2020two}, which provide fixed, non-programmable entropy generation but lack algorithmic versatility.  

\clearpage

\clearpage

\begin{table*}[ht!]
\centering
    \caption{Comparison of FPGA implementation of PRNG.} 
    \begin{tabular}{ccccccccc} 
         \hline
         PRNG &   \makecell[cc]{Area Metrics \\ (LUT*, FF, DSP)}&  \makecell[cc]{Frequency \\(MHz)}   
&\makecell[cc]{Power \\(mW)} &\makecell[cc]{Normalized \\ Power \\ (mW/MHz)}&  \makecell[cc]{Energy  \\ per bit \\ (nJ/bit)}
&\makecell[cc]{Throughput \\(bit/clock)}   & \makecell[cc]{Targeted \\Hardware}& \makecell[cc]{NIST \\ Test}  \\
         \hline
         LCG-Based \cite{Panda2020} &  (440, 128, 0)  &  282.64  &36.84 &0.13&  0.13&1   & \makecell[cc]{Xilinx \\ Vertix-7}& passed \\
         \hline         
         BBS-based \cite{Panda2020} &  (539, 181, 0)  &  161.41  &24.53&0.15&  10.48&1/69   & \makecell[cc]{Xilinx \\ Vertix-7}& passed \\
         \hline         
         Chaos-based \cite{Silva2023} &  (242, 65, 8) &  61.94  &114.00&1.87&  -&-  & \makecell[cc]{Xilinx \\ Artix-7}& passed \\
         \hline 
        cNN-based \cite{GarzonGonzalez2025}& (12215, 3117, 40)&  \makecell[cc]{0.867 \\ (effective)} &- & -&  - &8 & \makecell[cc]{Altera  \\ Cyclone V \\ SoC}&passed\\
         \hline   
         NPRNG &  (4544, 3441, 0) &  6.00 &3.26 &0.54&  27.16&$\approx$1/50  & \makecell[cc]{Lattice \\ Semiconductor \\  iCE40UP5K}& passed \\
         \hline  
    \end{tabular}
    * The LUTs in Xlinix FPGAs (Vertix-7 and Artix-7) have 5 inputs, while iCE40UP5K and Cyclone V SoC have  4-input LUTs.
    \label{table_comparison}
\end{table*}

\section*{Discussion}

The ability to generate bit streams that are indistinguishable from truly random bit streams usually involves trade-offs.  Some existing PRNGs will produce poor pseudo-random bit streams \cite{randu}, while others require substantial computational resources to implement in hardware \cite{bbs2}.  Here, we show that asymptotically provably random spiking neural networks connected in a balanced configuration produce high-quality random bit streams that pass all 15 tests for the NIST SP800-22 package at similar or higher rates than other PRNGs. The closest PRNG class to our method is based on chaotic Neural Networks (cNNs) which combine chaos theory and artificial neural networks \cite{Anusree2025}. As one of the simplest cNNs, discrete Hopfield Neural Networks (HNNs) comprising two homogeneous \cite{Bao2025} or heterogeneous \cite{Bao2024} neurons are shown to be able to generate hyperchaotic attractors. Recently, continuous Memristive Hopfield Neural Networks (MHNNs) have been widely studied and used for security applications \cite{Lin2025,Yu2023}. MHNNs with three neurons \cite{Yu2023} and two neurons \cite{Lin2025} can generate multi-scroll chaotic attractors that serve as pseudo-random number generators for XOR-based medical image encryption. To prototype these complex systems, Field-Programmable Gate Arrays (FPGAs) have been widely used due to their reconfigurability and parallel processing capabilities \cite{GarzonGonzalez2025}.

The NPRNG described here is readily implemented in low-power FPGA hardware, using as little as 3.2 mW of power in simulation and 5.45 mW of power in fabricated hardware to generate a bit stream at a rate of 120 Kbits per second.  NPRNGs also have a key advantage over other PRNGs: the number of unique NPRNGs grows geometrically with network size.  For a network of $N$ neurons, O($2^{N^2}$) unique configurable NPRNGs can be produced.   We also remark that with the current hardware implementation using 5.45 mW, NPRNGs can be wireless powered (for example through NFC). One potential area of improvement for neuromorphic based pseudo-random number generation is the low-bandwidth ($\approx$120 kbps).    Some existing chaos based PRNGs achieve higher bandwidths in both FPGA/ARM implementations \cite{chaos_map_1,chaos_map_2}.  We expect that translating the NPRNGs considered here into  standard-cell ASIC nodes or dedicated neuromorphic processors will increase the bandwidth of an NPRNG, while keeping the power consumption in the sub milliwatts range.   

To evaluate the security profile of this architecture precisely, we explicitly describe its performance across four domains: (1) Statistical Randomness, which is verified empirically by the successful passage of the 15-test NIST SP-800-22 battery across both software and FPGA implementations; (2) Operational Unpredictability, driven by the structural sensitivity of the deterministic balanced chaotic state trajectory to localized single-spike or single-weight perturbations; (3) State Secrecy, the many-to-one output transformation may obscure the relationship between the observable bitstream and internal network state, although resistance to state-reconstruction, side-channel, and cryptanalytic attacks remains to be established; and (4) Cryptographic Security. We note that a formal proof of cryptographic security against arbitrary bounded adversaries falls outside the scope of this work. However, the high dimensionality and the inherent parallel hardware architecture of NPRNGs yields an important advantage compared to conventional PRNGs regarding its security; for instance the traditional solutions such as advanced encryption standard (AES)\cite{aes1,aes2}  may be vulnerable to side-channel analysis because secret-dependent computation can leak through power and electromagnetic emissions. By contrast, the NPRNGs have a more complex key using $O(N^2)$ bits (recurrent weights) and because the physical weights are localized and integrated with parallel Tree-Adder data paths, inferring individual weight states from cumulative dynamic power/electromagnetic signatures presents a computational hurdle. While a formal evaluation against side-channel or fault exploitation remains a subject for future work, this configurable, high-dimensional, parallel architecture of NPRNGs provides a low-power statistical random number generator applicable for neuromorphic edge computing.

From a broader engineering perspective, the primary application of this architecture lies in its capacity for a native subcomponent within neuromorphic processing units, providing sources of statistical entropy to execute tasks such as stochastic computing and probabilistic inference. By relying entirely on native spiking primitives, our NPRNG eliminates the clock-domain and routing overhead associated with traditional, non-neural PRNG macros. Furthermore, this balanced recurrent SNN core offers an efficient dual-utility framework: the same block of digital logic can operate as a low-power statistical entropy engine or be dynamically reconfigured as an reservoir computing accelerator for temporal data processing \cite{nicola1}. This flexibility delivers a highly integrated, reconfigurable building block optimized for resource-constrained edge computing environments. 

The most immediate follow-up on our work would be the construction of neuromorphic true random number generators (NTRNGs).  True random number generators are an ongoing topic of study due to their widespread use in cryptography \cite{truerng1,truerng2}.  Although there are many instantiations of a putative NTRNG, a straightforward implementation would be of an analog spiking network built on basic electrical circuit elements: resistors, capacitors, and recently fabricated memristors \cite{chua1,chua2,reram1} that would implement the gates of the voltage-gated ion channels involved in generating action potentials in neurons.  We remark that there are other TRNGs that can also achieve low energy cost per bit, from 0.3 picojoules per bit to 2 nanojoules per bit \cite{qtrng1,qtrng2}.  The low energy required may also be achievable in an NPRNG with an analog ASIC design. Moreover, to demonstrate the power efficiency of NPRNG as a neuromorphic model, the digital design can be converted to an analog or a mixed-signal implementation using emerging technologies such as near-memory or in-memory computing.  Other potential applications in a fully or partially design include SNN-based physically unclonable functions (PUFs).

Collectively, our results demonstrate the feasibility of NPRNGs as low power random number generators that use only 5.45 mW of power to generate pseudo-random bit streams at 120kbps in a small form factor.  We anticipate that NPRNGs will have applications in edge computing, where the low power and area constraints necessitate novel solutions. 

\section*{Acknowledgments (not compulsory)}
WN is funded by an NSERC Discovery Grant, a Canada Research Chair, a Hotchkiss Brain Institute start-up grant and the Cumming Medical Research Fund.  AG is supported by NSERC, Digital Research Alliance of Canada, Hotchkiss Brain Institute, Alberta Children's Hospital Research Institute, and the Azrieli Accelerator. This research was partially funded by Synaptrain Technologies Inc.

\section*{Author contributions statement}
WN and AG supervised the project.  WN, NA, and JS performed numerical simulations.  JS designed the FPGA/hardware.  WN, AG, NA, JS, and SS prepared and edited the manuscript. 

\section*{Conflict of Interest Statement}
WN, AG, and SS are the CSO, CPO, and CEO of Synaptrain Technologies Inc., respectively.   

\section*{Code Availability Statement} 
Code can be found at https://modeldb.science/2018267 with referee access password rx114.  Code will be made public upon acceptance of manuscript.

\clearpage
\section*{Methods}

\subsection*{Spiking Neural Networks and the Excitatory/Inhibitory Balanced State}

The network equations for a leaky-integrate-and-fire (LIF) network of spiking neurons in the excitatory/inhibitory balanced regime are given by
\begin{eqnarray}
 \tau_m \dot{v}_i(t) &=& -v_i + \sum_{j=1}^N w_{ij}r_j(t)+I \label{v}\\
 \dot{r}_j(t) &=&  -\frac{r_j}{\tau_d} + h_j\label{r} \\
 \dot{h}_j(t)&=& -\frac{h_j}{\tau_r} + \frac{1}{\tau_d\tau_r}\sum_{t<t_{jk}}\delta(t-t_{jk}) \label{h}
\end{eqnarray}
The voltage dynamics are given by equation (\ref{v}) where $v_i(t)$ is the voltage of the $i$th neuron.  When the neuron reaches a threshold, $v_i(t^-) = v_{th}$, it is said to fire a spike, and is reset to $v(t^+)= v_{reset}$.  The membrane time constant $\tau_m$ controls how much of the past history of currents arriving to neuron $i$ is filtered in between spikes.  The parameter $I$ acts as a bias current to each neuron and sets the background firing rate of the neurons without any synaptic inputs. Every spike fired by a neuron is filtered by a double exponential synaptic filter, $r(t)$, implemented by equations (\ref{r})-(\ref{h}) \cite{ermentrout}.   Briefly, this acts to filter the $k$th spike fired by the $j$th neuron at time $t_{jk}$ with a synaptic current pulse.  This pulse has a rise time given by $\tau_r$ and a decay time given by $\tau_d$.  Similar networks were considered previously in \cite{nicola1,nicola2}.  The function $\delta(t)$ is the Dirac delta function.  

The weight matrix $w_{ij}$ couples the $j$th neuron to the $i$th neuron and is randomly generated to elicit high-dimensional chaotic dynamics state in both recurrent rate and recurrent spiking neural networks \cite{ei1,ei2,F1,F2,F3,B1,B2,B3,harish,S1}.  In particular, the conditions on the moments of $w_{ij}$ that may elicit this state are 
\begin{eqnarray}
E(w_{ij}) &=& 0 \label{r1} \\
E(w_{ij}^2) &=& \frac{g^2}{N} \label{r2}
\end{eqnarray}
where $g$ is sufficiently large.  In spiking neural networks, the chaos displayed can occur in two forms depending on the $g$ parameter: rate chaos or Poisson-like irregular spiking \cite{S1,harish}. For larger $g$, the network enters into the rate chaos regime, with the specific transition point determinable with dynamic mean-field theories 
\cite{harish,S1}. For smaller $g$, the spikes having Poisson-like firing statistics, where the rate is homogeneous both across the network, and in time.   Note that the irregular dynamics displayed in the ``spike-chaos" are irregular, but exhibit a kind of stable chaos \cite{F3,nicola1}.  This arises as perturbations to the voltages of the neurons may or may not cause a lasting change to an irregular spiking trajectory 
\cite{F3,F1}.  In analytical systems that allow for the construction of exact, event-based spike-to-spike simulations, the discrete spikes are shown to be chaotic \cite{F1} with the number of positive Lyapunov exponents scaling linearly with the network size $N$ \cite{F1}.  

To satisfy the constraints in equations (\ref{r1})-(\ref{r2}), while simultaneously allowing for simple implementations in neuromorphic hardware, we use bipolar values of the weights:
\begin{eqnarray}
w_{ij} = \pm\frac{g}{\sqrt{N}}.
\end{eqnarray}
Further, each neuron $i$ receives exactly $N/2$ positive weights ($+g/\sqrt{N}$) and $N/2$ negative weights ($-g/\sqrt{N})$, which implies that equation (\ref{r1}) holds exactly for each neuron $i$.

\subsection*{Generating Uniform Random Bit Sequences from a Spiking Neural Network with a Dynamic Lookup Table}

Lookup tables are commonly used tools in computer science \cite{aes,des}.  A dynamic lookup table was used to resolve the RNG quality issues that arise with the relative refractory period of spikes as follows.  Let $n$ denote the index of the $n$th spike fired by the network.  Let $y_n$ be defined as:
\begin{eqnarray}
y_n = \mod(y_{n-1}+s_n,N)
\end{eqnarray}
where $N$ is the modulus operation and $s_n$ is the index of the neuron that fired the $n$th spike.  The variable $y_n$ increments the indices of the $n$ spikes fired modulo $N$.  Then, the $n$th bit reported by the network is given by 
\begin{eqnarray}
b_n = L(y_n)
\end{eqnarray}
where
$$ L(i) = b_i $$
and $\bm b$ is a randomly generated $N\times 1$ binary vector containing $N/2$ bits that are 0, and $N/2$ bits that are 1, randomly permuted. 

\subsection*{The NIST-SP-800-22 Package for Testing PRNGs}

NIST SP 800-22 is a set of statistical tests designed to evaluate the randomness of binary sequences generated by random number generators.  NIST outlines a suite of tests to ensure that a PRNG produces sequences that exhibit statistics that cannot be distinguished from those derived from a truly random binary sequence\cite{NIST}. 

NIST SP 800-22 includes 15 tests, each designed to evaluate different aspects of randomness in a binary sequence (a bit stream). The tests look for patterns or biases that could indicate the sequence is not random, and thus the null hypothesis for all of the tests below is that the bit stream is generated from independent and identical Bernoulli random variables.  Here, we summarize the tests briefly, with greater details found in \cite{NIST}.  Note that in all cases below, a sequence ``passing a NIST test" implies having a p-value above (rather than below) a critical value. For certain tests, such as Non-overlapping Template Matching Test, Random Excursions Test, and Random Excursions Variant Test, multiple results were reported due to variations in test parameters and the nature of the tests themselves. To maintain a rigid evaluation criterion, we adopted a strict passing condition for these tests: a test was considered passed only if all reported results for that test met the passing threshold. Specific parameters for all of the tests are listed in Table \ref{tab:parameters}.

\begin{table}[ht]
\centering
\begin{tabular}{|l|l|}
\hline
\textbf{Parameter} & \textbf{Value} \\
\hline
Input length (length of a bit stream) & 1,000,000 \\
\hline
Number of bitstreams & 100 \\
\hline
Block Frequency Test - block length (M) & 128 \\
\hline
Non Overlapping Template Test - block length (m) & 9 \\
\hline
Overlapping Template Test - block length (m) & 9 \\
\hline
Approximate Entropy Test - block length (m) & 10 \\
\hline
Serial Test - block length (m) & 16 \\
\hline
Linear Complexity Test - block length (M) & 500 \\
\hline
\end{tabular}
\caption{\label{tab:parameters} Parameters used for all NIST experiments}
\end{table}

\subsection*{Microarchitecture of the NPRNGs}
Since the network equations, (\ref{v}), (\ref{r}), and (\ref{h}), are translated into the main computation blocks in the microarchitecture, their complexity has a significant impact on the space or time complexity of the corresponding hardware blocks and the overall microarchitecture. Simplifying the multiply-accumulate (MAC) operations, as the fundamental but hardware-costly blocks of neural networks, will lead to efficient implementation of the microarchitecture. Our hardware architecture achieves a multiplier-less footprint by completely algorithm-hardware co-design, developing the network constraints alongside the digital datapath. Specifically, the original equations were modified by moving multiplications in (\ref{v}) to (\ref{h}) due to the linearity of the synaptic equations. This resulted in converting the synaptic multiplication to the addition operation in which the synaptic weights are added in the occurrence of spikes. This is an immediate benefit of using spikes for processing. The discretized equations are:
\begin{eqnarray}
v_i[n+1] &=&  v_i[n] + \frac{\Delta}{\tau_m} \left(-v_i[n] + r_i[n] + I \right),\label{vd} \\
r_i[n+1]  &=&  r_i[n] + \Delta \left(-\frac{1}{\tau_d}  r_i[n] + h_i[n] \right),\label{rd} \\
h_i[n+1]  &=&  h_i[n] + \Delta \left(-\frac{1}{\tau_r} h_i[n] + \frac{1}{\tau_r\tau_d} \sum_{j=1}^N w_{ij} \delta[n-n_{jk}]  \right),\label{hd} \\
y[n+1] &=&  mod(i_{spk} + y[n],N)\label{fd}
\end{eqnarray}
where $n$ and $\Delta$ are the time instance and duration of each time step.

Figure. \ref{figure3}A shows the microarchitecture of the PRNG comprising a data path unit and a control unit. The data path unit includes a memory for the synaptic weights  ($w_{ij}$) and the computational blocks. To initialize the network in the first time step, the seed with the size of $N$ bits is applied to the $v[n]$ block where the $j$th neuron's initial condition is set to super-threshold (spiking) if the $j$th bit of an input seed is 1.  If the $j$th bit of the input seed is 0, the initial condition for the neuron is $v_i(t)=v_{reset}$. 

During the operational phase, the synaptic weights are multiplied by the spikes. The synaptic weights are organized into a $N \times N$ memory, each location comprising one bit. To save memory space, one bit is used for each weight and the CONVERT  block converts the binary values ($0$ and $1$) to the signed values ($-1$ and $+1$). Moreover, the multiplication of the synaptic weights to the spikes does not require a multiplier, but a multiplexer (MUX) because of the nature of spikes.  $N$ CONVERT and MUX blocks are used in parallel that correspond to $N$ weights per neuron. The synaptic multiplication results are passed to the adder tree that calculates the term $\sum_{j=1}^N w_{ij} \delta[n-n_{jk}]$ in (\ref{hd}). An adder tree has a parallel computing structure that efficiently computes the sum of multiple numbers by recursively adding pairs of numbers until a final sum is obtained.

The blocks of $h[n], r[n], v[n]$ and $y[n]$ implement the corresponding equations of \ref{hd}, \ref{rd}, \ref{vd}, and \ref{fd}, respectively. To reduce the hardware area, resource sharing was used in which each of these blocks was used in a time-multiplexed fashion for calculations of all synapses and neurons \cite{Shamsi2024}. The resource sharing reduces throughput by the order of $N$ - the number of neurons. To compensate for throughput, the pipeline design was used to implement the synaptic and neuron's dynamics. Fig. \ref{figure3}A (right) shows the pipeline design of the synaptic and neuron's dynamics ($h[n]$, $r[n]$, and $v[n]$). The pipeline architecture breaks the combinational circuits into smaller parts to increase the throughput of the system at the cost of a small increase in total circuit latency and consumption of a few registers \cite{Shamsi2024} (Supplementary Section S3). While the number of stages in $h[n]$ block is two, the number of stages in the $r[n]$ block and $v[n]$ is taken equally (five). A 32-bit fixed-point representation with 16 bits for the integer part and 16 bits for the fraction part were used in the arithmetic calculations. As it is shown in Fig. \ref{figure3}A, the multiplications in blocks $h[n]$, $r[n]$, and $v[n]$ were implemented using shift and add operations. Since all of the constant parameters are of the form $\pm 2^k$ where $k$ is an integer, they are represented precisely. The truncation method was employed for rounding purposes in the intermediate calculations.

At each clock cycle, the calculation for each block is computed and fed into the next block. The values of $h[n], r[n]$ and $v[n]$ are stored in the Random Access Memory (RAM) to be used in the next time-step. The spikes in each clock cycle are also stored in a serial-in-parallel-out (SIPO) register to be used in the next time step. The index of the neurons generating spikes are sent to block $y[n]$, which is implemented by a $log_2(N)$-bit unsigned adder. Finally, the $y[n]$ values are converted to output bits through the $N \times 1$ look-up table (L). 

The control unit controls the flow of computations in the data path unit, which is conceptually based on a modular block, counter module (CM), comprising a counter and two comparators \cite{Shamsi2024}. The counter is enabled through a pulse with the duration of at least one clock cycle at the ON pin and stays enabled until it receives a pulse at the OFF pin in which it is disabled and the counter value resets to zero. In this block, the counter is always disabled when the counter value reaches $N-1$. The EN pin is $1$ when the counter is enabled. 

Since the read and write operations of the RAMs in the synaptic and neuron blocks are performed sequentially, the output of the CM1 block is used to generate the read address (W-R-add) of the weight memory, and it is also used as a reference block to trigger the other CM blocks. The A, B, C and D are related to the number of pipeline stages in the $h[n]$ block ($H_{stage}$), $r[n]$ and $v[n]$ block ($RV_{stage}$),and adder tree ($A_{stage}$): 
$
A = A_{stage}-2,
B = A_{stage} + H_{stage},
C = A_{stage} + H_{stage}-2,
D = A_{stage} + H_{stage} + RV_{stage}.
$
Note that the constant value of $-2$ is used, since in FPGAs it usually takes two clock cycles after providing the address for the RAM memory to make the content of the memory available.

The space complexity of the microarchitecture was determined by the growth of the hardware resource usage when the number of neurons $N$ increases. The time complexity was considered as increasing the number of clocks per time-step of the SNN integration, $\Delta$. Table \ref{table_complexity} summarizes the space and time complexity for each block. The synaptic memory with the size of $N \times N$ stores the weights and its space complexity is $\mathcal O(N^2)$. Since the weights are organized as $N \times N$ and $N$ clocks are needed to read the weights in each time-step, the time complexity is $\mathcal O(N)$. The total number of CONVERT and MUX blocks is $N$ resulting in the space complexity of $\mathcal O(N)$. Since these blocks are parallel combinational circuits, the calculations of each neuron is done in one clock cycle. To perform all calculations in each time step, $N$ clocks are required, resulting in the time complexity is $\mathcal O(N)$. Given $N$ inputs, the number of stages and the total number of adders in the pipeline adder tree are $[log_2(N)]$ and $[log_2(N)]+1$, respectively. Therefore, the space complexity of the adder is $\mathcal O(log(N))$, while the time complexity is $\mathcal O(N)$. Although the blocks of $h[n], r[n], v[n]$ are shared for all synapses and neurons in the network, they contain a RAM memory with a size of $N$, therefore their space and time complexities are $\mathcal O(N)$ and $\mathcal O(N)$, respectively. The L block with the size of $N \times 1$ has the space of $\mathcal O(N)$. Since the maximum number of spikes in each time step can be $N$, therefore the time complexity of reading the bits from look-up table is $\mathcal O(N)$. Considering the maximum space and time complexity of the blocks, the overall complexity of space and time of the microarchitecture were considered as $\mathcal O(N^2)$ and $\mathcal O(N)$, respectively.
\begin{table*}[ht]
\centering
\caption{The space and time complexity of the blocks in the microarchitecture.}
\begin{tabular}[t]{lcclccccccc}
\hline
Complexity& Memory&CONVERT& MUX&Adder-Tree& $h[n]$& $r[n]$& $v[n]$& $y[n]$& $L[y]$&Overall (Max)\\
\hline
Space&$\mathcal O(N^2)$&$\mathcal O(N)$&$\mathcal O(N)$&$\mathcal O(log(N))$& $\mathcal O(N)$& $\mathcal O(N)$& $\mathcal O(N)$&$\mathcal O(1)$&$\mathcal O(N)$&$\mathcal O(N^2)$\\
Time&$\mathcal O(N)$&$\mathcal O(N)$&$\mathcal O(N)$&$\mathcal O(N)$&$\mathcal O(N)$& $\mathcal O(N)$& $\mathcal O(N)$&$\mathcal O(N)$&$\mathcal O(N)$&$\mathcal O(N)$\\
\hline
\end{tabular}
\label{table_complexity}
\end{table*}%

\subsection*{Additional Methods for Each Figure} 
\subsubsection*{Figure 1}
The network of spiking neurons in Figure 1A-F was integrated with a forward Euler integration scheme with a time step of $\Delta = 0.05$ ms.  The parameters can be found in Table \ref{tp}.  The network was simulated with a total period of 100 seconds with random initial conditions where the voltage was uniformly distributed in the interval $[v_{reset},30]$ mV.   The row sum of the weight matrix was precisely balanced (to 0) with the procedure described in Supplementary Section S1.  The interspike-intervals were estimated as 
$$ ISI_{jk} =  t_{jk} - t_{jk-1}$$
where $t_{jk}$ is the $k$th spike fired by the $j$th neuron.  The ISI distribution was computed with a histogram with a bin size of 0.001 seconds.

The coefficient of variation for a neuron was defined as:
$$
CV_j = \frac{\sigma_{ISI_j}}{\mu_{ISI_j}}
$$

where 
$$ \mu_{ISI_j} = \frac{1}{n_j-1}\sum_{k=2}^{n_j} {ISI_{jk}}$$
is the mean ISI for neuron $j$ which fired $n_j$ spikes and 
$$ \sigma_{ISI_j} = \sqrt{\frac{1}{n_j-1}\sum_{k=2}^{n_j} \left(ISI_{jk}-\mu_{ISI_j}\right)^2} $$
is the standard deviation of the ISI's for neuron $j$. 

In the spike failure/deletion experiment (Figure \ref{figure1}E), the network was first simulated with an initial seed without deletion.  Then a single spike was prevented from eliciting a post-synaptic response by manually deleting the increment in the synaptic filter that would otherwise occur.  In the weight-flip experiment, the initially positive $w_{21}$ was set to $-w_{21}$ after $t=2$ seconds for the remainder of the simulation. 

For the quantized network (Figure \ref{figure1}G-J), the network was simulated identically to Figure 1, only with dimensionless quantized parameters (Table \ref{tp}).  The integration time-step was also quantized with $dt = 2^{-10}$.   The network was initialized with the identical uniform distribution in $v_i(t)$ as in the non-quantized case, and with all other variables as ($h_i(t), r_i(t)$) at 0.

\begin{table}[htp!]
\centering
\begin{tabular}{| c|c|c|}
  \hline			
  $N$ & 256 & 256  \\
   \hline		
  $\tau_m$ & 10 ms & $2^{-7}$ \\
   \hline	
  $v_{reset}$ & -65 mV & 0 \\
   \hline	
     $v_{peak}$ & -40 mV & $2^4-1$ \\
 \hline
   $\tau_D$ & 20 ms &$2^{-6}$   \\
   \hline	
     $\tau_R$ & 2 ms & $2^{-9}$ \\
   \hline	
  $g$ & 0.1 &$2^{-5}$  \\
   \hline	
\end{tabular}
\caption{Table of parameters for spiking neural network simulations.}\label{tp}
\end{table}

\subsubsection*{Figure 2}

The networks in Figure \ref{figure2}D were simulated with identical parameters as the quantized network in Figure 1.   The network was simulated for a total of $T=3$ time units (dimensionless).  The sinusoidal input provided to initialize the network was 
$$ c(t) = \sin(10\pi t) $$
for $t<=1$.  The input $c(t)$ is set to 0 for $t>1$. 
The randomly generated step function consisted of random positive or negative deflection for $t=0.05$ time units, followed by 0 deflection at time for $0.05$ time units, periodically.  These deflections persisted for $t<1$ and were shut off after.  

The networks in Figure \ref{figure2}E-H contained identical parameters as in Figure 2D, however the synchronization signals used to initialize a network were randomly generated with LFSR (See supplementary Methods).  Each network was simulated until 115 million spikes were fired, with the last 100 million spikes used in the NIST testing.

\subsubsection*{Supplementary Figure 1}  
The parameters for the neurons and network are identical as in Figure 1 (non-quantized network with binary weights), with a total simulation time of 100 seconds.  The networks were initialized identically to Figure 1.  The ISI distribution/Coefficient of Variation were identically computed as in Figure 1.  The log of the histogram count was plotted for clarity in Supplementary Figure S1-S2.   

\subsubsection*{Supplementary Figure 2}
The simulations in Supplementary Figure 2 are identical to Supplementary Figure 1, only with the quantized network parameters from Figure 1. 

\subsubsection*{Supplementary Figure 3}
The networks were quantized and simulated for a total of 5 time units, with identical parameters as in Figure 1.  The input signal to these networks was on in the interval $[0.5,0.75]$ (Supplementary Figure S3A)
 or $[0.5,1.25]$ (Supplementary Figure S3B).  The synchronization signal was defined as 
 $$ c(t) = 40\text{sign}(\cos(10\pi t))$$

\subsubsection*{Supplementary Figure 4}

The quantized network had identical parameters to Figure 1.  The network was simulated for 100 time units, at which point the cross-correlation between neurons $i$ and $j$ were computed with the MATLAB \emph{crosscorr} function with a time lag of 100 time steps, corresponding to approximately 0.1 time units for neurons 1 to 5.

\subsubsection*{Supplementary Figure 5}

The average number of NIST tests passed for the networks simulated in Figure \ref{figure2}G-H was computed and correlated against the average coefficient of variation.  The average for each simulation was computed as the average CV over all neurons:

$$ \bar{\mu}_{CV} = \frac{1}{256}\sum_{j=1}^{256} CV_{j}$$

A least squares model was fit to the average number of NIST tests passed as a function of the average CV with the MATLAB \emph{polyfit} function.   For each parameter set, we varied the connection strengths $g_q = 2^{-j}$ for $j=1,2,\ldots 16$ and varying synaptic time constants $\tau_D = 2^{-i}$, for $i=1,2,\ldots 8$. Using the same random seed for each configuration, we performed 1000 simulations. The system was simulated until it produced the required number of bits/spikes ($10^8$) for NIST testing.  

\subsubsection*{Supplementary Figure 6}
The quantized network was simulated with identical parameters as in Figure 1.  The weight matrix was precisely balanced, but non-randomly generated with a banding structure along the diagonal.  In particular, for neuron 1, the first $N/2$ weights were positive while the last $N/2$ weights were negative ($\pm g/\sqrt{N}$).  Each subsequent neuron in the network had identical weights as in neuron 1, but with a circular shift operation applied to shift the row matrix 1 entry to the left. The network was simulated for a total of 0.4 time units.

\subsubsection*{Supplementary Figure 7}
The maximum Lyapunov exponent was computed numerically using the MATLAB subfunction \emph{lyapunovExponent}, which computes the Lypaunov exponent with a sampled time-series by using a delay embedding.   The delay parameter was taken to be 1 second, while 20 second simulations were used for each network considered.  The networks considered were both quantized and binarized.

\clearpage

\clearpage
\section*{Supplementary Information}

\section*{Supplementary Videos}
\textbf{Supplementary Video 1}:  Simulation of an excitatory/inhibitory balanced network for 0.3 seconds of simulation time.  (Left) $N=256$ Neurons aligned on a $16\times16$ grid.  Pixels switching from black to white indicates a spike was fired.   (Middle) The spike raster-plot for the corresponding simulation for the $N=256$ neurons.  (Right) The voltage traces for 5 neurons.  

\section*{Supplementary Section S1:  Parameter Complexity Space} 

For a network of $N$ neurons with binary weights, there are $N^2$ weights, each with a possible value of 
\begin{eqnarray}
w_{ij} = \pm \frac{g}{\sqrt{N}}
\end{eqnarray}
This implies $2^{N^2}$ possible configurations of the weights, in the case where each weight is a Bernoulli random variable with possible configurations $+g/\sqrt{N}$ and $-g/\sqrt{N}$. 

However, for the NPRNGs considered here, each row of the weight matrix has exactly $\frac{N}{2}$ positive values and $\frac{N}{2}$ negative values, with every row being a permutation of the vector 
$$ w_0 = \frac{g}{\sqrt{N}}[1,1,1,\ldots 1,-1,-1,\ldots -1]$$ 
where the first $N/2$ entries are $\frac{g}{\sqrt{N}}$ and the last $N/2$ entries are $-\frac{g}{\sqrt{N}}$.
For a single row, there are $C(N,\frac{N}{2})$ unique configurations of $\bm w_0$, where $C(N,k) = \frac{N!}{(N-k)!k!}$.  As there are $N$ rows in $\bm w$, there are 
$$ N_w = C\left(N,\frac{N}{2}\right)^N = \left(\frac{N!}{((N/2)!)^2}\right)^N $$
possible weights.    The limiting behaviour of $N_w$ can be computed with Stirling's asymptotic formula:
$$ \log_2(n!) = n\log_2 n - n\log_2 (e) + O(\log_2(n))$$
with 
\begin{eqnarray}
\log_2(N_w) &=& N\left(\log_2\left(N!\right)-2\log_2\left(\left(\frac{N}{2}\right)!\right)\right)\\
&=& N \left( N\log_2 N - N\log_2 (e) + O(\log_2(N)) -  N\log_2 N/2 + N\log_2 (e) - O(\log_2(N/2))
\right)\\
&=&  N \left( N(\log_2 N - \log_2(N/2)) + O(\log_2(N))) 
\right)\\
&=& N \left(N - O(\log_2(N))\right)\\
&=& N^2 - O(N\log_2(N))
\end{eqnarray}
and thus 
\begin{eqnarray}
N_w \sim 2^{N^2 - O(N\log_2(N))} \quad N\rightarrow \infty 
\end{eqnarray}
Thus, for large $N$ we should expect that the number of weight matrices $N_w$ satisfying the constant-row sum condition scales like $2^{N^2}$, which effectively is identical to the scaling of not having any row-sum constraint.  

\section*{Supplementary Section S2:  Simulations}  

To conduct the extensive simulations, we began by ensuring that the software implementation was consistent with the hardware. This was validated by running both the software and hardware using the same set of weights and seed values, then comparing their spike raster plots, which showed perfect alignment. Once confirmed, the software simulation was used to generate random numbers for the experiments. In total, approximately $10^8$ random bits were generated, divided into 100 bit streams of $10^6$ bits each. This division ensured compatibility with the NIST testing package, which requires at least $10^6$ bits per bit stream for some statistical tests. All experiments were executed using the Slurm \cite{slurm} job scheduler on the Digital Research Alliance of Canada clusters: Beluga, Cedar, Graham, and Narval. According to the Digital Alliance, the total CPU usage for our experiments was equivalent to 65.49 core years, where a core year represents continuous computation on a single CPU core for one year.

For running each test, there are two separate phases. In the first phase, $10^8$ random bits are generated using a different set of network weights, a look-up table, and a seed value. To ensure the generation of diverse random numbers, the network must be initialized with different seeds. While a variety of methods can be employed to generate seeds for the network, we used a 16-bit Linear Feedback Shift Register (LFSR) as a straightforward and effective approach to initialize the network with $2^{16}$ distinct configurations. The characteristic polynomial of the LFSR is as follows:

\begin{eqnarray}
P(x) = x^{16} + x^{15} + x^{13} + x^{4} + 1
\end{eqnarray}

Once the network is initialized and has run for a specified duration, we transition to the second phase: capturing the random bits. To standardize output across experiments, we define a file size threshold of $115 MB$. The SNN, configured with different parameters, produces random bits at varying rates, and the simulation stops once this predefined size is achieved.

In the second phase, the generated random bits were evaluated for their randomness using the NIST SP 800-22 testing package, as described in the Methods section. The results of the NIST tests were collected for each set of experiments and processed using Python scripts. First, the raw reports were parsed into CSV files for easier analysis. Final results were then generated by applying a strict evaluation metric: a test was considered "passed" only if all instances of that test met the minimum pass rate criteria for both the P-value and Proportion metrics as specified in the NIST documentation. This ensured that only experiments meeting the most stringent randomness requirements were reported as successful.

\section*{Supplementary Section S3:  Throughput of the microarchitecture}  

The advantage of using pipeline design as a temporal parallelism is to reduce the propagation delay associated with the critical path in a combinational  circuit. This results in increasing the maximum clock frequency and throughput (more operations completed per time unit). In our case, we define throughput as the number of neurons processed per time:
\begin{eqnarray}
TH = \frac{N_{proc}}{T_{proc}}
\end{eqnarray}
where $N_{proc}$ and $T_{proc}$ are the number of neurons processed and the processing time, respectively.
When $T_{proc}=\Delta$ (one time step), $N_{proc} = 256$, therefore:
\begin{eqnarray}
TH = \frac{256}{\Delta} = \frac{256}{T_{clk}.N_{clk}}
\end{eqnarray}
where $T_{clk}$ and $N_{clk}$ are the clock period and the number of clocks per time step. 

Without a pipeline design, the minimum value of $N_{clk}$ can be 256. The value of $T_{clk}$ is determined by the critical path delay $T_{cp}$.  Thus, the throughput without the pipeline design is:
\begin{eqnarray}
TH_{no-pip} = \frac{256}{256.T_{cp}} = \frac{1}{T_{cp}}
\end{eqnarray}
With a pipeline design,  the number of neurons processed is $N_{proc} = 256$, while the clock cycles will increase to $N_{clk} = 256+S$, where $S$ is the number of pipeline stages added to break the critical path into smaller segments. Note that in the design considered here, the last neuron must be updated before starting the new time step. That is why we added the latency $S$ to the number of clocks per time step.
In the pipeline design, the clock period is determined by the maximum stage delay  $T_{stg}$. If we assume equal stage delays for simplicity, the following approximation holds:
\begin{eqnarray}
T_{stg} = \frac{T_{cp}}{S} + S.T_{reg}
\end{eqnarray}
where $T_{reg}$ is the overhead of the pipeline registers, which is usually much smaller than the critical path delay, $T_{cp}$. Thus, the throughput with the pipeline design is:
\begin{eqnarray}
TH_{pip} = \frac{256}{T_{clk}.N_{clk}}= \frac{256}{(256+S).(T_{cp}/S+S.T_{reg})} 
\end{eqnarray}
The pipeline design increases throughput under the conditions $TH_{pip}>TH_{no-pip}$, resulting in:
\begin{eqnarray}
T_{cp}(256.S-256)&>& (256+S).S.T_{reg}
\label{TH}
\end{eqnarray}
As $S$ (the number of pipeline stages) grows, the left side of \ref{TH} increases linearly with $S.T{cp}$, and the right side grows roughly $S^2.T_{reg}$. There is an intermediate sweet spot where enough stages reduce the critical path delay, while the register overhead does not dominate.

\section*{Supplementary Section S4:   Expected Cycle Time}  

In this section, we analyze how long before a cycle would be expected in performing a discrete time, discrete space simulation of the balanced chaotic spiking neural network considered here.  Each neuron consists of $3$ variables:  $(v_i(t),r_i(t),h_i(t))$ which are discretized with 32 bits of precision.  The size of the discrete state-space, $S(N)$ is then given by 
$$ S(N) = 2^{D\times n_{bits} \times N}= 2^{24576}$$ 
for the parameters considered here ($N=256$, $n_{bits} = 32$, $D = 3$).  This sets the maximum time step before we are guaranteed to observe a cycle by the pigeon-hole principle.  However, the birthday paradox sets a lower time before the probability of detecting any cycle is dramatically increased. This occurs at approximately $\approx \sqrt{S(N)} = 2^{12288}$.   We remark that the birthday party heuristic is inline with recent evaluations of the cycle time of the Lorenz96 dynamical system, which is a spatiotemporal system consisting of $N$ coupled chaotic Lorenz systems \cite{cycle1}   It was shown numerically that cycles in the Lorenz96 system increase in period exponentially with the system size.

\clearpage
\section*{Supplementary Figures} 
\begin{figure}[htp!]
\centering
    \includegraphics[scale=0.6]{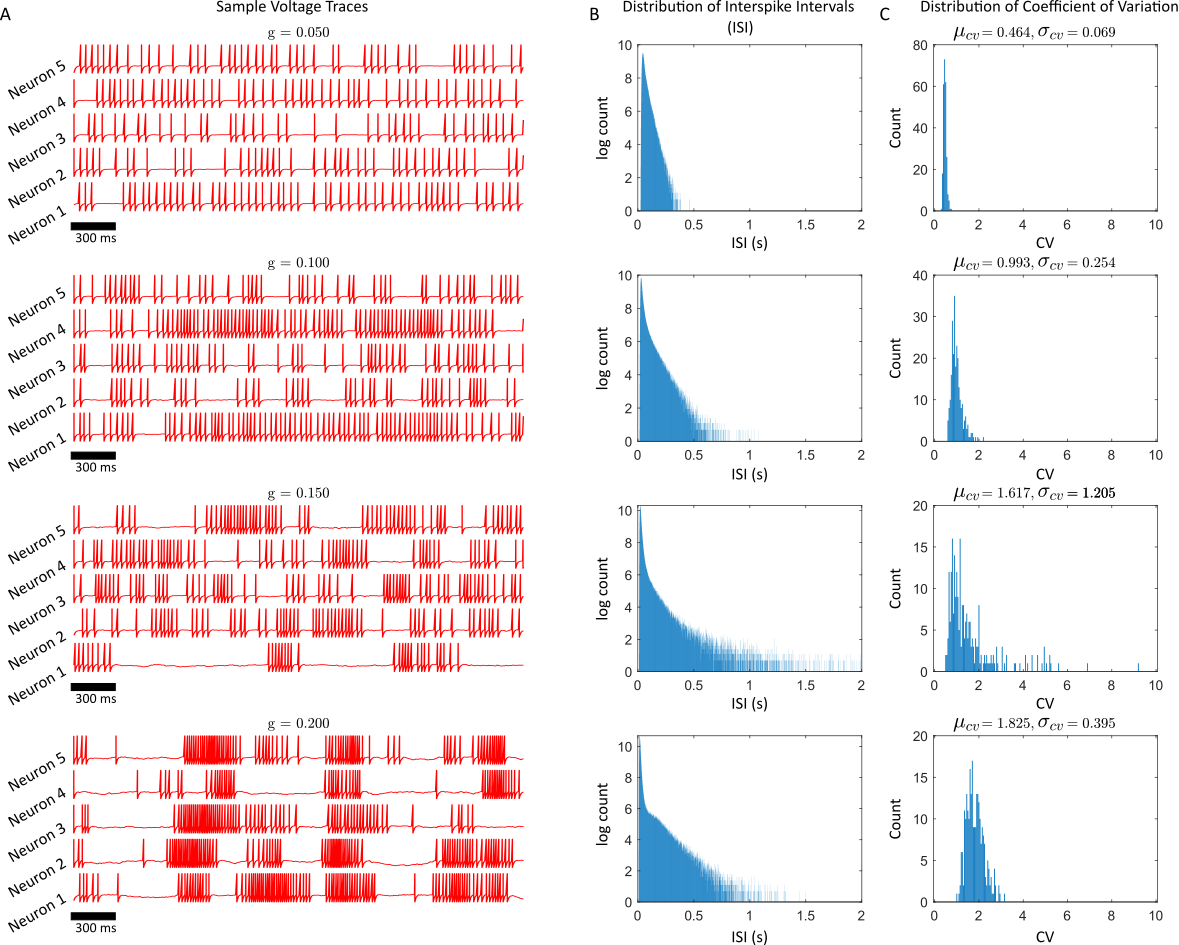}
    \caption*{\textbf{Supplementary Figure S1}: The behavior of the non-quantized SNN with binarized weights for varying connection strengths  
\textbf{(A)} The voltage traces for 5 neurons from four networks with increasing $g$ for a 3 second period.  Note that all 4 networks have the same weight matrix, only scaled by the different $g$ value ($w_{ij} = \pm \frac{g}{\sqrt{N}}$).  As $g$ is increased, the neurons fire bursts of varying size/duration.   Each network contains $N=256$ neurons. 
\textbf{(B)} The distribution of inter-spike-intervals for the 4 simulated networks from (A).  For lower values of $g$, the distributions resemble a Poisson process with a refractory period.   Note that the log-ISI count is used on the y-axis.
\textbf{(C)}  The distribution of the coefficient of variation for the networks in (A)-(B).  The mean and standard deviation of the CVs are shown in the title.  For more strongly coupled networks $g=0.15, g=0.2$, the mean CV is greater than 1, indicating the burstiness of the neurons in these coupling regimes.  }
\end{figure}

\clearpage

\begin{figure}[htp!]
\centering
    \includegraphics[scale=0.6]{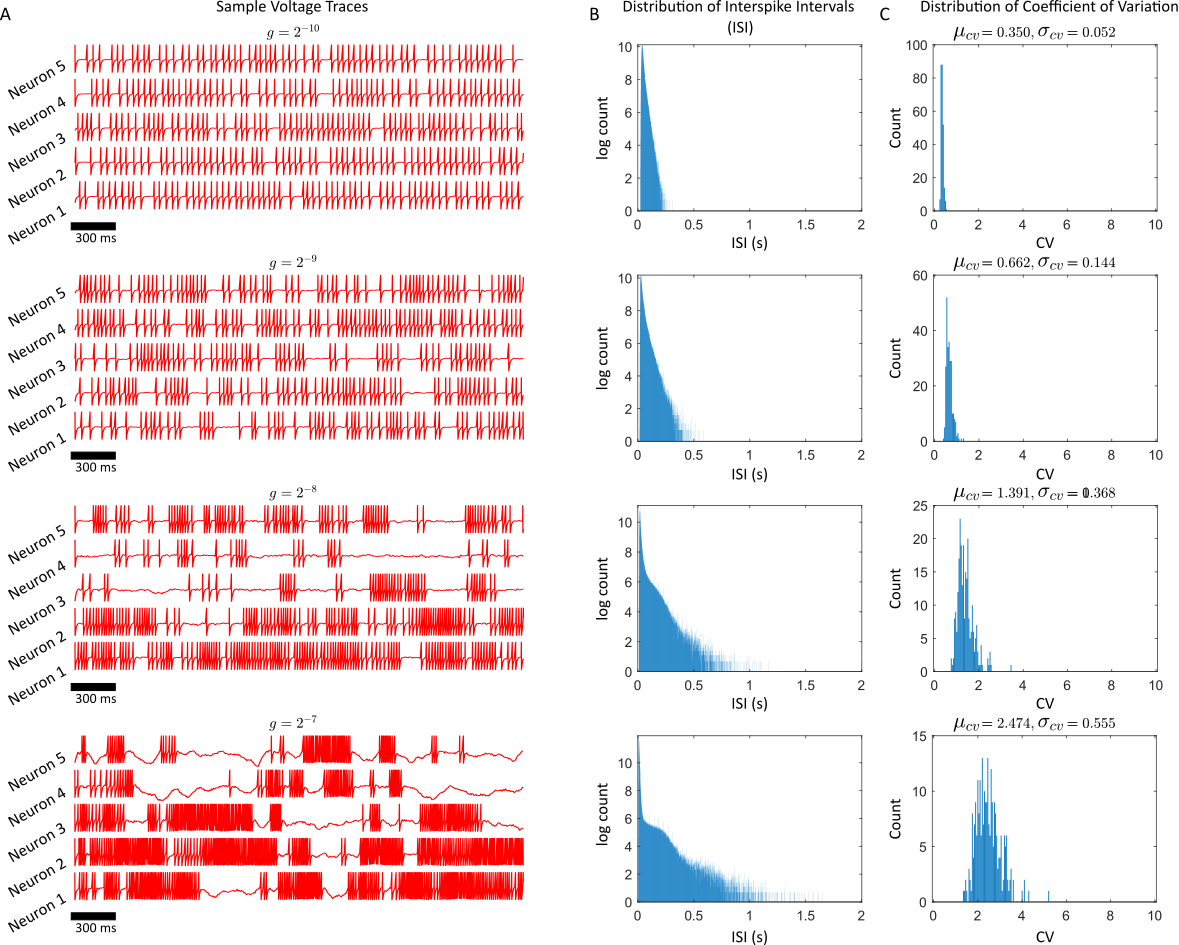}
    \caption*{\textbf{Supplementary Figure S2}: The behavior of a quantized SNN with binarized weights for varying connection strengths. \textbf{(A)} The voltage traces for 5 neurons from four networks with increasing quantized connection strength $g_{q} = 2^{-n}$ for a 3 second period.  Note that all 4 networks have the same weight matrix, only scaled by the different $g_q$ value. As $g_q$ is increased, the neurons fire bursts of varying size/duration.   Each network contains $N=256$ neurons.  All operations in the network are quantized.
\textbf{(B)} The distribution of inter-spike-intervals for the 4 simulated networks from (A).  For lower values of $g_q$, the distributions resemble a Poisson process with a refractory period.   Note that the log-ISI count is used on the y-axis.
\textbf{(C)}  The distribution of the coefficient of variation for the networks in (A)-(B).  The mean and standard deviation of the CVs are shown in the title.  For more strongly coupled networks $g_q=2^{-8}, g_q=2^{-7}$, the mean CV is greater than 1, indicating the burstiness of the neurons in these coupling regimes. 
}
\end{figure}

\clearpage

\begin{figure}[htp!]
\centering
    \includegraphics[scale=0.45]{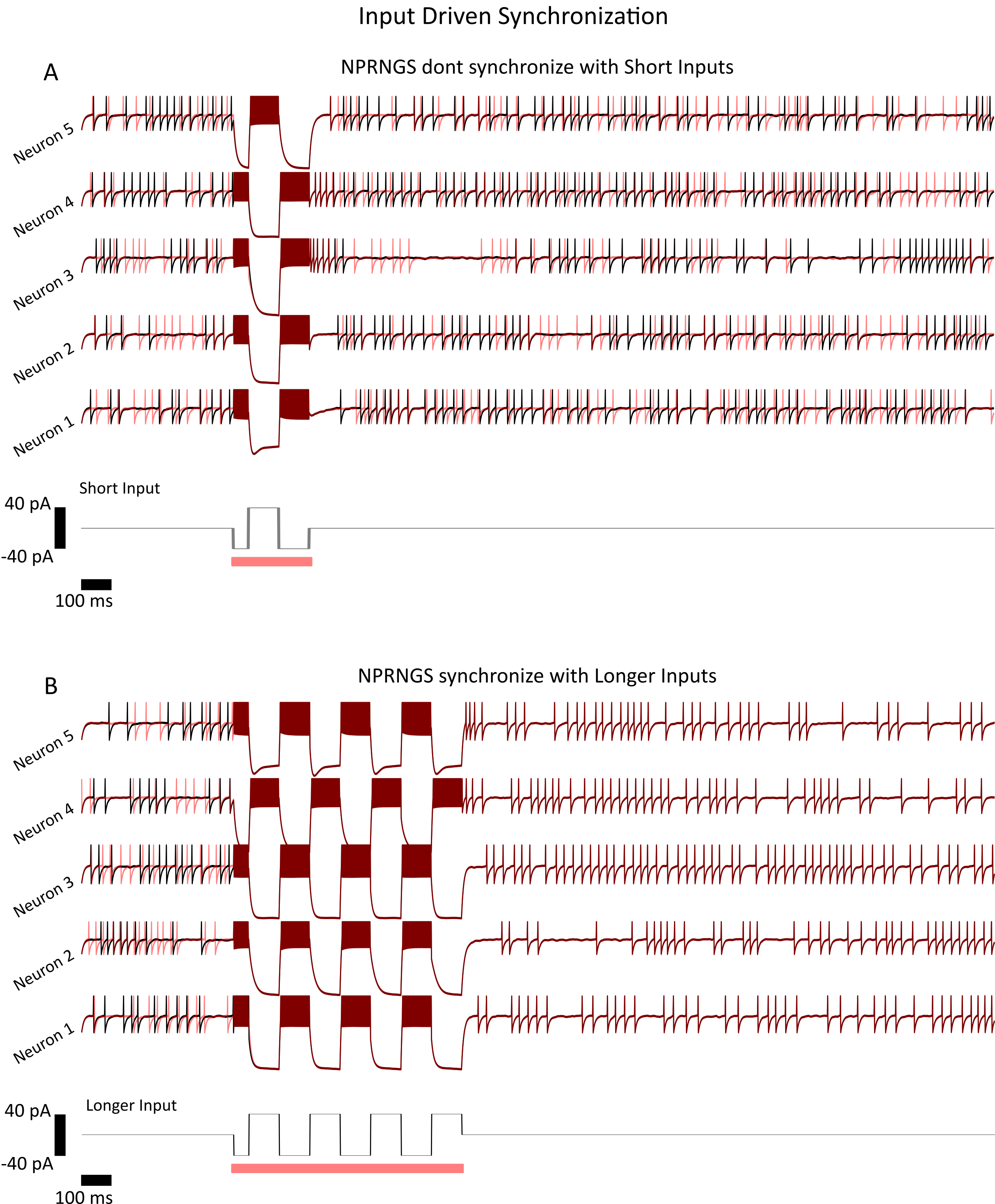}
    \caption*{\textbf{Supplementary Figure S3}: Synchronizing NPRNGs with input currents.  \textbf{(A)}  The voltage traces for 5 neurons, of the same balanced, and quantized network of $N=256$ neurons.  Voltages for each neuron are initialized at different values for both simulations (red, black).  An input current is applied to each neuron, multiplied by an input weight (Materials and Methods). The input current is non-zero for a short duration of time (250 ms)  The input current causes a transient synchronization of the neurons in the network.  Eventually, the neuron's desynchronize.  \textbf{(B)}  Identical to (A), only the input current is applied for a longer duration of time (750 ms).  The neurons maintain synchrony, despite the chaotic dynamics, in perpetuity.}
\end{figure}

\clearpage

\begin{figure}[htp!]
\centering
    \includegraphics[scale=0.45]{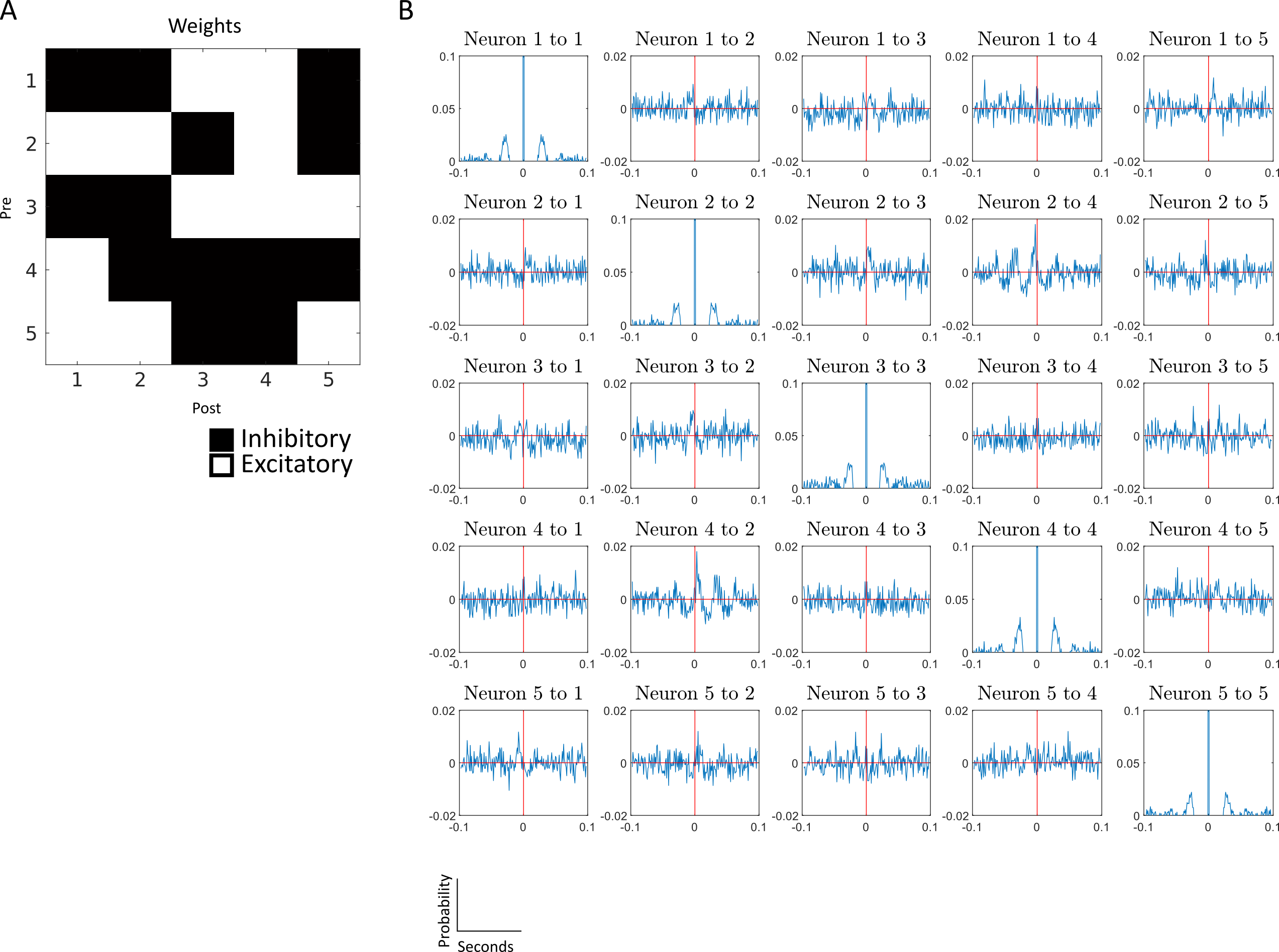}
    \caption*{\textbf{Supplementary Figure S4}: Correlation structure in a Balanced-Quantized Network. \textbf{(A)}  The sign of the weights between 5 neurons in a network of $N=256$ neurons with balanced connectivity, and quantized operations.  \textbf{(B)} The autocorrelation (diagonal) and cross-correlations (off-diagonal) between neurons in a balanced, quantized network simulated for $T=1000$ seconds.  The cross-correlations are weak, and scale-like $N^{-1}$.  The autocorrelation functions have prominent peaks in between a refractory period of non-firing centered at 0.}
\end{figure}
\clearpage

\begin{figure}[htp!]
\centering
    \includegraphics[scale=0.4]{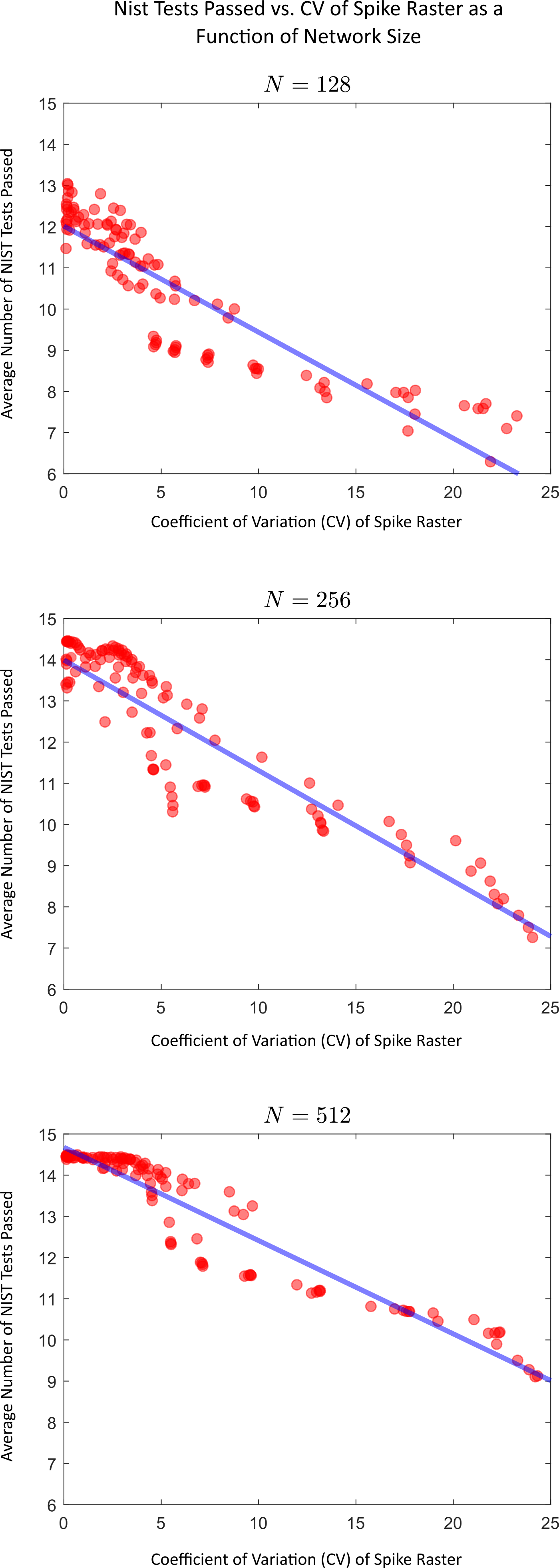}
    \caption*{\textbf{Supplementary Figure S5}: Average CV versus the average number of NIST tests passed.  The mean CV is inversely correlated with the quality of random numbers generated ($\rho = -0.9029$, $p\ll 10^{-4}$, $N=128$, $\rho = -0.6818$, $p\ll 10^{-4}$, $N=256$, $\rho = -0.8147$, $p\ll 10^-4$, $N=128$, $\rho = -0.9538$, $p\ll 10^{-4}$, $N=512$).  The slops and intercepts for the line of best fits are -0.2576, -0.2684, -0.2256, and 12.0134, 13.9926, and 14.6775, for $N=128$, $N=256$, and $N=512$ respectively. }
\end{figure}
\clearpage

\begin{figure}[htp!]
\centering
    \includegraphics[scale=0.6]{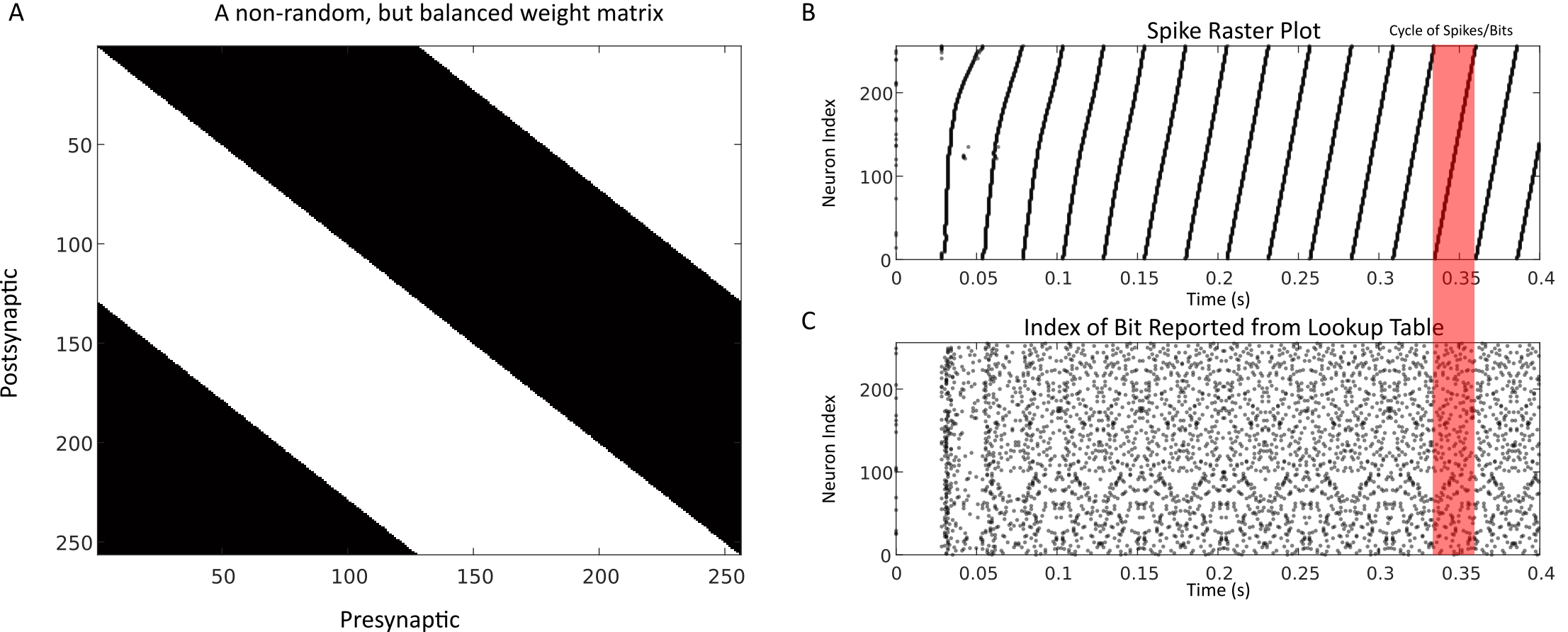}
    \caption*{\textbf{Supplementary Figure S6}: Poorly generated but still balanced excitatory/inhibitory synaptic weights lead to cycling.  \textbf{(A)} The weight matrix is generated with 50\% excitatory and 50\% inhibitory connections. Each row is a circular shift of the previous row, one index to the right. \textbf{(B)} The spike raster plot for a simulated quantized network of neurons coupled with the weights in (A).  The network eventually stabilizes on a fixed periodic sequence of spiking.  \textbf{(C)} The index of the bit reported in the look-up table.  The bits reported in this case also cycle, but at twice the period of the neurons in the network.}
\end{figure}
\clearpage

\begin{figure}[htp!]
\centering
    \includegraphics[scale=0.7]{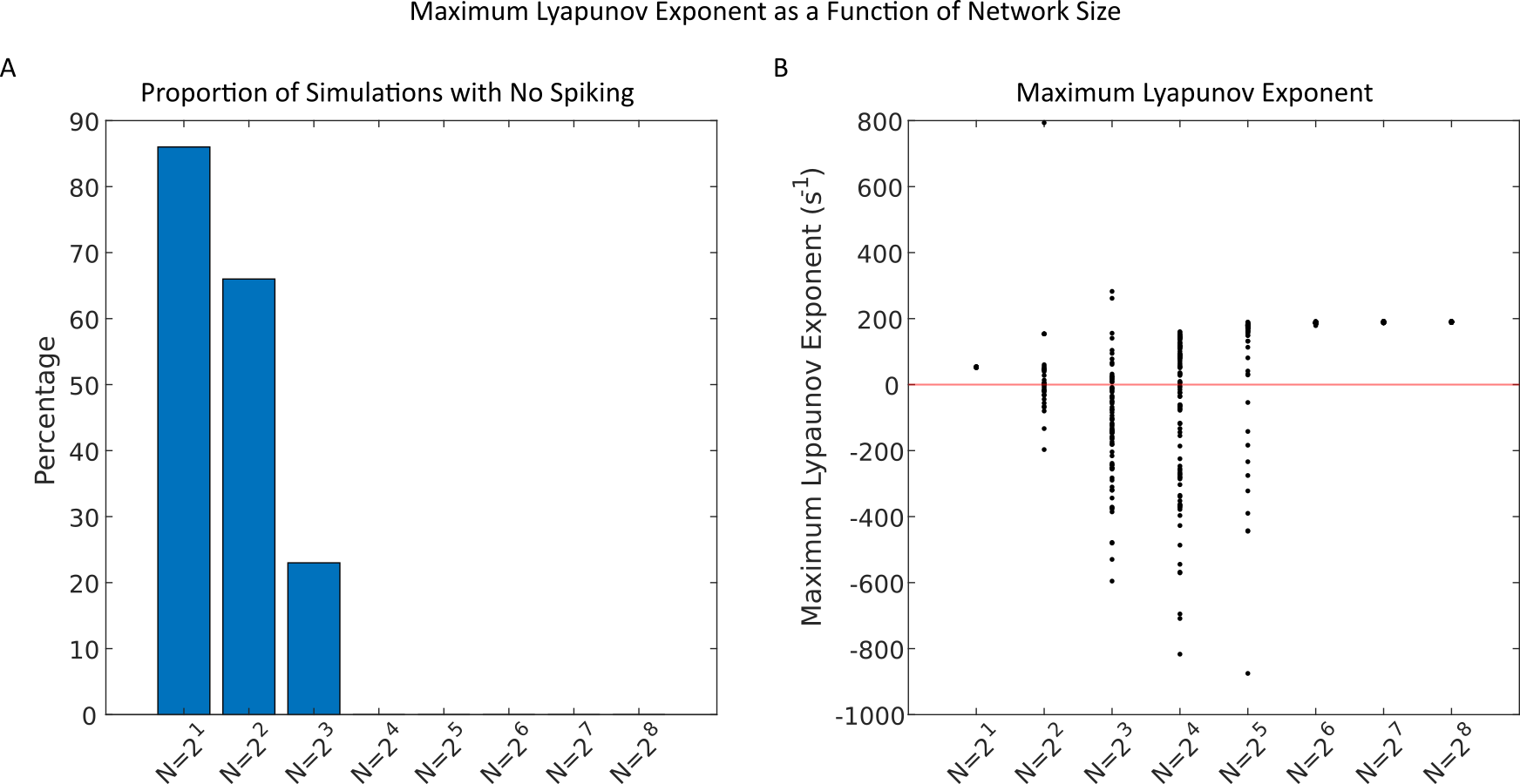}
    \caption*{\textbf{Supplementary Figure S7}: The maximum Lyapunov exponent as a function of $N$, the network size, for increasingly larger $N$. The coupling parameter was scaled as $g = \frac{2^{-6}}{\sqrt{N}}$.  (Left) The proportion of simulations where the spiking eventually ceased, as the non-spiking state is bistable with other attractors.  (Right) The numerically computed maximum Lyapunov exponent as a function of $N$.  The red line denotes the 0 point. }
\end{figure}
\clearpage

\begin{figure}[htp!]
\centering
    \includegraphics[scale=0.5]{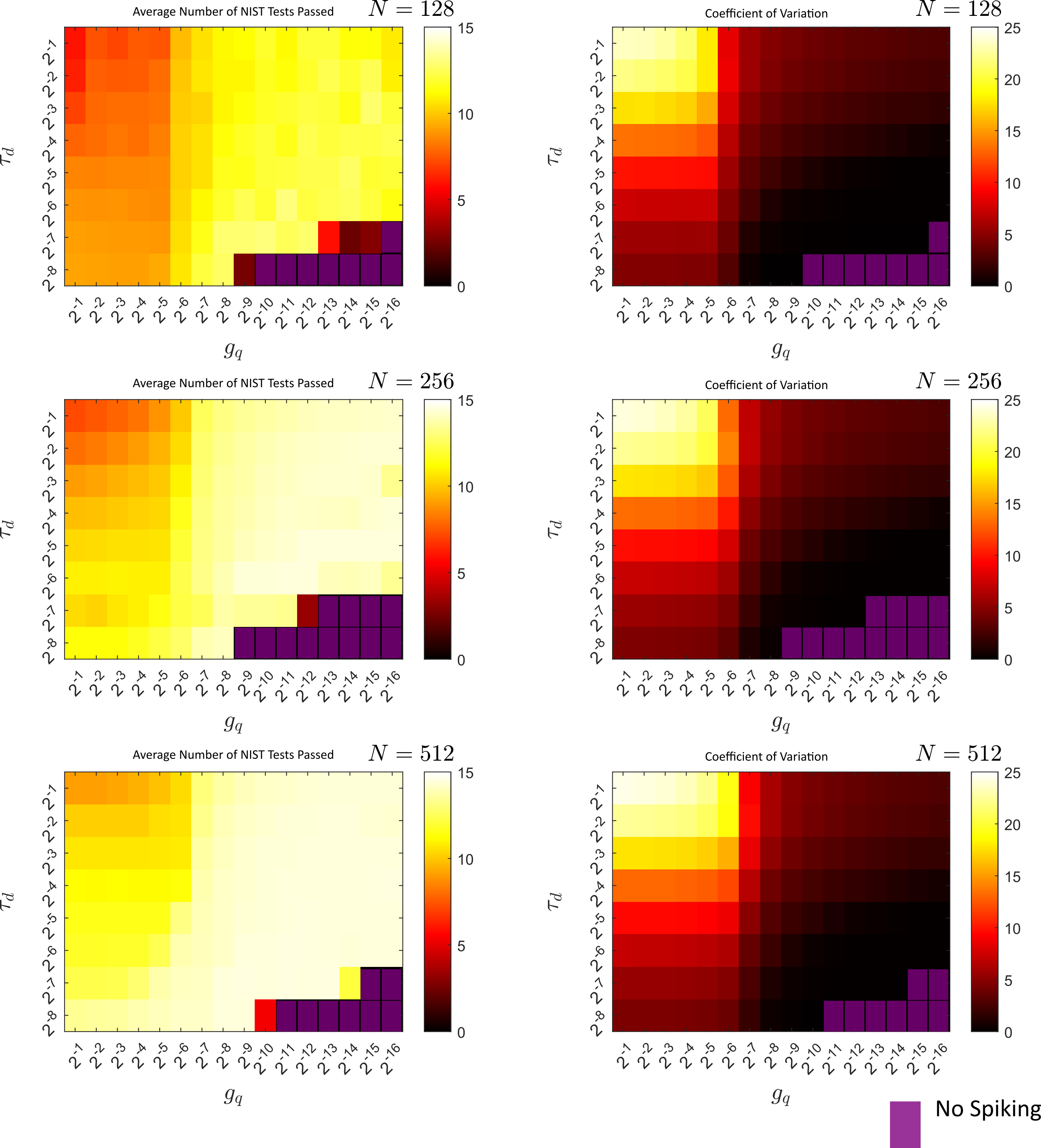}
    \caption*{\textbf{Supplementary Figure S8}: Heat map of the average number of NIST tests passed (left) and the coefficient of variations of the resulting spike trains (right) for $N=128$ (top row), $N=256$ (middle row), and $N=512$ (bottom row).    }
\end{figure}
\clearpage

\begin{figure}[htp!]
\centering
    \includegraphics[scale=0.45]{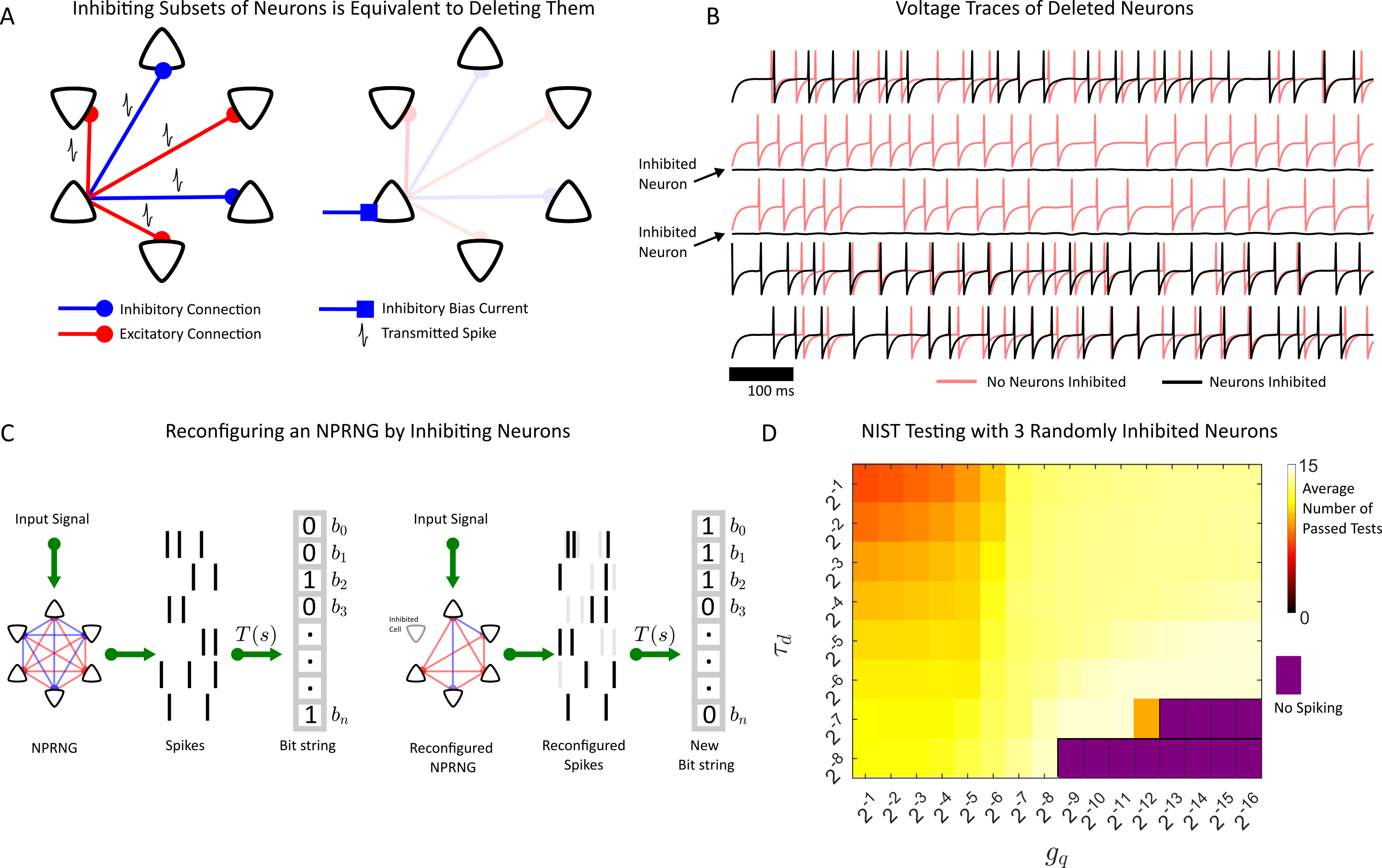}
    \caption*{\textbf{Supplementary Figure S9}: \textbf{(A)} Schematic of inhibition of neurons via bias currents.  An externally inhibited neuron (blue) no longer transmit spikes in the network.  This is equivalent to setting all of the post-synaptic weights for the inhibited inside the weight matrix to 0. This generates a novel NPRNG from an existing NPRNG. \textbf{(B)} A simulation of the binarized/quantized spiking neural network with 2 neurons inhibited (black) and no inhibitory currents (red).  The neurons are inhibited by providing a sub-threshold bias current ($I=0$).  The spike-raster of the two networks is different. \textbf{(C)} Neuronal inhibition allows and chaotic dynamics allows for multiple NPRNGs to be generated from a single network.  \textbf{(D)} NIST testing of networks of $N=256$ neurons with $N=3$ neurons randomly inhibited for the duration of each simulation.  Deleting neurons does not appreciably change PRNG quality.}
\end{figure}
\clearpage

\bibliography{REFs}

@misc{Anusree2025,
   author = {M. Anusree and P. Nair Pramod},
   doi = {10.1007/s11071-025-11503-7},
   issn = {1573269X},
   journal = {Nonlinear Dynamics},
   publisher = {Springer Science and Business Media B.V.},
   title = {Understanding chaotic neural networks: A comprehensive review},
   year = {2025}
}

@article{Bao2025,
   author = {Han Bao and Minqi Xi and Haigang Tang and Xi Zhang and Quan Xu and Bocheng Bao},
   doi = {10.1109/TII.2024.3488800},
   issn = {1551-3203},
   issue = {2},
   journal = {IEEE Transactions on Industrial Informatics},
   month = {2},
   pages = {1862-1871},
   publisher = {IEEE Computer Society},
   title = {Discrete Two-Heterogeneous-Neuron HNN and Chaos-Based Hardware Poisson Encoder},
   volume = {21},
   url = {https://ieeexplore.ieee.org/document/10756211/},
   year = {2025}
}

@article{Bao2024,
   author = {Bocheng Bao and Haigang Tang and Yuanhui Su and Han Bao and Mo Chen and Quan Xu},
   doi = {10.1109/TCSI.2024.3382259},
   issn = {15580806},
   issue = {12},
   journal = {IEEE Transactions on Circuits and Systems I: Regular Papers},
   pages = {5907-5918},
   publisher = {Institute of Electrical and Electronics Engineers Inc.},
   title = {Two-Dimensional Discrete Bi-Neuron Hopfield Neural Network With Polyhedral Hyperchaos},
   volume = {71},
   year = {2024}
}

@article{Yu2023,
   author = {Fei Yu and Hui Shen and Qiulin Yu and Xinxin Kong and Pradip Kumar Sharma and Shuo Cai},
   doi = {10.1109/TNSE.2022.3223930},
   issn = {2327-4697},
   issue = {2},
   journal = {IEEE Transactions on Network Science and Engineering},
   month = {3},
   pages = {845-858},
   publisher = {IEEE Computer Society},
   title = {Privacy Protection of Medical Data Based on Multi-Scroll Memristive Hopfield Neural Network},
   volume = {10},
   url = {https://ieeexplore.ieee.org/document/9960712/},
   year = {2023}
}

@article{Lin2025,
   author = {Hairong Lin and Xiaoheng Deng and Fei Yu and Yichuang Sun},
   doi = {10.1109/TCAD.2024.3429410},
   issn = {19374151},
   issue = {1},
   journal = {IEEE Transactions on Computer-Aided Design of Integrated Circuits and Systems},
   pages = {304-316},
   publisher = {Institute of Electrical and Electronics Engineers Inc.},
   title = {Diversified Butterfly Attractors of Memristive HNN with Two Memristive Systems and Application in IoMT for Privacy Protection},
   volume = {44},
   year = {2025}
}

@article{GarzonGonzalez2025,
   author = {Jeferson Andres Garzon Gonzalez and Jose de JesusRangel-Magdaleno and Jesus Manuel Munoz-Pacheco},
   doi = {10.1109/TII.2024.3523548},
   issn = {19410050},
   issue = {4},
   journal = {IEEE Transactions on Industrial Informatics},
   pages = {3097-3106},
   publisher = {IEEE Computer Society},
   title = {FPGA Implementation of a Multi-PRNG Based on a Multiscroll Chaotic Hopfield Neural Network},
   volume = {21},
   year = {2025}
}

@article{Panda2020,
   author = {Amit Kumar Panda and Kailash Chandra Ray},
   doi = {10.1109/TIM.2019.2909248},
   issn = {0018-9456},
   issue = {4},
   journal = {IEEE Transactions on Instrumentation and Measurement},
   month = {4},
   pages = {1011-1019},
   publisher = {Institute of Electrical and Electronics Engineers Inc.},
   title = {A Coupled Variable Input LCG Method and its VLSI Architecture for Pseudorandom Bit Generation},
   volume = {69},
   url = {https://ieeexplore.ieee.org/document/8682119/},
   year = {2020},
}

@article{Silva2023,
   author = {Samuel Souza da Silva and Matheus Cardoso and Lucas Nardo and Erivelton Nepomuceno and Michael Hubner and Janier Arias-Garcia},
   doi = {10.1109/TIM.2023.3235457},
   issn = {0018-9456},
   journal = {IEEE Transactions on Instrumentation and Measurement},
   pages = {1-8},
   publisher = {Institute of Electrical and Electronics Engineers Inc.},
   title = {A New Chaos-Based PRNG Hardware Architecture Using the HUB Fixed-Point Format},
   volume = {72},
   url = {https://ieeexplore.ieee.org/document/10013689/},
   year = {2023},
}

@article{Shamsi2024,
   author = {Jafar Shamsi and Wilten Nicola},
   doi = {10.1109/TCSI.2024.3463536},
   issn = {1549-8328},
   journal = {IEEE Transactions on Circuits and Systems I: Regular Papers},
   pages = {1-11},
   publisher = {Institute of Electrical and Electronics Engineers Inc.},
   title = {Implementation of Linear Differential Equations Using Pulse-Coupled Oscillators With an Ultra-Low Power Neuromorphic Realization},
   url = {https://ieeexplore.ieee.org/document/10697267/},
   year = {2024},
}

@inproceedings{Cardoso2021,
   author = {Matheus B.R. Cardoso and Samuel S. da Silva and Lucas G. Nardo and Roberto M. Passos and Erivelton G. Nepomuceno and Janier Arias-Garcia},
   doi = {10.1109/ISCAS51556.2021.9401653},
   isbn = {9781728192017},
   issn = {02714310},
   booktitle = {Proceedings - IEEE International Symposium on Circuits and Systems},
   publisher = {Institute of Electrical and Electronics Engineers Inc.},
   title = {A new PRNG hardware architecture based on an exponential chaotic map},
   volume = {2021-May},
   year = {2021},
}

@article{Kopparthi2022,
   author = {Venkata Reddy Kopparthi and Anil Kali and Samrat L. Sabat and Kiran Kumar Anumandla and Rangababu Peesapati and J.S. Armand Eyebe Fouda},
   doi = {10.1016/j.aeue.2022.154138},
   issn = {14348411},
   journal = {AEU - International Journal of Electronics and Communications},
   month = {4},
   pages = {154138},
   publisher = {Elsevier GmbH},
   title = {Hardware architecture of a digital piecewise linear chaotic map with perturbation for pseudorandom number generation},
   volume = {147},
   url = {https://linkinghub.elsevier.com/retrieve/pii/S1434841122000358},
   year = {2022},
}

@article{Frenkel2023,
   author = {Charlotte Frenkel and David Bol and Giacomo Indiveri},
   doi = {10.1109/JPROC.2023.3273520},
   issn = {0018-9219},
   issue = {6},
   journal = {Proceedings of the IEEE},
   month = {6},
   pages = {623-652},
   title = {Bottom-Up and Top-Down Approaches for the Design of Neuromorphic Processing Systems: Tradeoffs and Synergies Between Natural and Artificial Intelligence},
   volume = {111},
   url = {https://ieeexplore.ieee.org/document/10144567/},
   year = {2023},
}

@article{Pedersen2024,
   author = {Jens E. Pedersen and Steven Abreu and Matthias Jobst and Gregor Lenz and Vittorio Fra and Felix Christian Bauer and Dylan Richard Muir and Peng Zhou and Bernhard Vogginger and Kade Heckel and Gianvito Urgese and Sadasivan Shankar and Terrence C. Stewart and Sadique Sheik and Jason K. Eshraghian},
   doi = {10.1038/s41467-024-52259-9},
   issn = {2041-1723},
   issue = {1},
   journal = {Nature Communications},
   month = {9},
   pages = {8122},
   pmid = {39285176},
   publisher = {Nature Research},
   title = {Neuromorphic intermediate representation: A unified instruction set for interoperable brain-inspired computing},
   volume = {15},
   url = {https://www.nature.com/articles/s41467-024-52259-9},
   year = {2024},
}

@article{Lazaro2020,
   author = {Antonio Lazaro and Marti Boada and Ramon Villarino and David Girbau},
   doi = {10.1109/ACCESS.2019.2962570},
   issn = {2169-3536},
   journal = {IEEE Access},
   pages = {2200-2221},
   publisher = {Institute of Electrical and Electronics Engineers Inc.},
   title = {Study on the Reading of Energy-Harvested Implanted NFC Tags Using Mobile Phones},
   volume = {8},
   url = {https://ieeexplore.ieee.org/document/8943414/},
   year = {2020},
}

@misc{NIST,
  author = {Lawrence Bassham and Andrew Rukhin and Juan Soto and James Nechvatal and Miles Smid and Stefan Leigh and M Levenson and M Vangel and Nathanael Heckert and D Banks},
  title = {A Statistical Test Suite for Random and Pseudorandom Number Generators for Cryptographic Applications},
  year = {2010},
  month = {2010-09-16},
  publisher = {Special Publication (NIST SP), National Institute of Standards and Technology, Gaithersburg, MD},
  url = {https://tsapps.nist.gov/publication/get_pdf.cfm?pub_id=906762},
  language = {en},
}

@article{ei1,
  title={Chaos in neuronal networks with balanced excitatory and inhibitory activity},
  author={Van Vreeswijk, Carl and Sompolinsky, Haim},
  journal={Science},
  volume={274},
  number={5293},
  pages={1724--1726},
  year={1996},
  publisher={American Association for the Advancement of Science}
}

@article{ei2,
  title={Chaos in random neural networks},
  author={Sompolinsky, Haim and Crisanti, Andrea and Sommers, Hans-Jurgen},
  journal={Physical review letters},
  volume={61},
  number={3},
  pages={259},
  year={1988},
  publisher={APS}
}

@article{B1,
  title={Slow dynamics and high variability in balanced cortical networks with clustered connections},
  author={Litwin-Kumar, Ashok and Doiron, Brent},
  journal={Nature neuroscience},
  volume={15},
  number={11},
  pages={1498--1505},
  year={2012},
  publisher={Nature Publishing Group US New York}
}

@article{B2,
  title={Balanced networks of spiking neurons with spatially dependent recurrent connections},
  author={Rosenbaum, Robert and Doiron, Brent},
  journal={Physical Review X},
  volume={4},
  number={2},
  pages={021039},
  year={2014},
  publisher={APS}
}

@article{B3,
  title={Balanced synaptic input shapes the correlation between neural spike trains},
  author={Litwin-Kumar, Ashok and Oswald, Anne-Marie M and Urban, Nathaniel N and Doiron, Brent},
  journal={PLoS computational biology},
  volume={7},
  number={12},
  pages={e1002305},
  year={2011},
  publisher={Public Library of Science San Francisco, USA}
}

@article{F1,
  title={Dynamical entropy production in spiking neuron networks in the balanced state},
  author={Monteforte, Michael and Wolf, Fred},
  journal={Physical review letters},
  volume={105},
  number={26},
  pages={268104},
  year={2010},
  publisher={APS}
}

@article{F2,
  title={Dynamical models of cortical circuits},
  author={Wolf, Fred and Engelken, Rainer and Puelma-Touzel, Maximilian and Weidinger, Juan Daniel Fl{\'o}rez and Neef, Andreas},
  journal={Current opinion in neurobiology},
  volume={25},
  pages={228--236},
  year={2014},
  publisher={Elsevier}
}

@article{F3,
  title={Dynamic flux tubes form reservoirs of stability in neuronal circuits},
  author={Monteforte, Michael and Wolf, Fred},
  journal={Physical Review X},
  volume={2},
  number={4},
  pages={041007},
  year={2012},
  publisher={APS}
}

@article{R1,
  title={Input correlations impede suppression of chaos and learning in balanced firing-rate networks},
  author={Engelken, Rainer and Ingrosso, Alessandro and Khajeh, Ramin and Goedeke, Sven and Abbott, LF},
  journal={PLOS Computational Biology},
  volume={18},
  number={12},
  pages={e1010590},
  year={2022},
  publisher={Public Library of Science San Francisco, CA USA}
}

@article{S1,
  title={Two types of asynchronous activity in networks of excitatory and inhibitory spiking neurons},
  author={Ostojic, Srdjan},
  journal={Nature neuroscience},
  volume={17},
  number={4},
  pages={594--600},
  year={2014},
  publisher={Springer US New York}
}

@article{S2,
  title={How connectivity, background activity, and synaptic properties shape the cross-correlation between spike trains},
  author={Ostojic, Srdjan and Brunel, Nicolas and Hakim, Vincent},
  journal={Journal of Neuroscience},
  volume={29},
  number={33},
  pages={10234--10253},
  year={2009},
  publisher={Soc Neuroscience}
}

@article{N1,
  title={Dynamics of sparsely connected networks of excitatory and inhibitory spiking neurons},
  author={Brunel, Nicolas},
  journal={Journal of computational neuroscience},
  volume={8},
  pages={183--208},
  year={2000},
  publisher={Springer}
}

@article{J1,
  title={Transition to chaos in random networks with cell-type-specific connectivity},
  author={Aljadeff, Johnatan and Stern, Merav and Sharpee, Tatyana},
  journal={Physical review letters},
  volume={114},
  number={8},
  pages={088101},
  year={2015},
  publisher={APS}
}

@article{Osc1,
  title={Transition from asynchronous to oscillatory dynamics in balanced spiking networks with instantaneous synapses},
  author={Di Volo, Matteo and Torcini, Alessandro},
  journal={Physical review letters},
  volume={121},
  number={12},
  pages={128301},
  year={2018},
  publisher={APS}
}

@article{destexhe1,
  title={Low-dimensional chaos in an instance of epilepsy.},
  author={Babloyantz, Agnessa and Destexhe, Alain},
  journal={Proceedings of the National Academy of Sciences},
  volume={83},
  number={10},
  pages={3513--3517},
  year={1986},
  publisher={National Acad Sciences}
}

@article{london1,
  title={Sensitivity to perturbations in vivo implies high noise and suggests rate coding in cortex},
  author={London, Michael and Roth, Arnd and Beeren, Lisa and H{\"a}usser, Michael and Latham, Peter E},
  journal={Nature},
  volume={466},
  number={7302},
  pages={123--127},
  year={2010},
  publisher={Nature Publishing Group UK London}
}

@article{harish,
  title={Asynchronous rate chaos in spiking neuronal circuits},
  author={Harish, Omri and Hansel, David},
  journal={PLoS computational biology},
  volume={11},
  number={7},
  pages={e1004266},
  year={2015},
  publisher={Public Library of Science San Francisco, CA USA}
}

@article{mh,
  title={Equation of state calculations by fast computing machines},
  author={Metropolis, Nicholas and Rosenbluth, Arianna W and Rosenbluth, Marshall N and Teller, Augusta H and Teller, Edward},
  journal={The journal of chemical physics},
  volume={21},
  number={6},
  pages={1087--1092},
  year={1953},
  publisher={American Institute of Physics}
}

@inproceedings{diffnets,
  title={Improved denoising diffusion probabilistic models},
  author={Nichol, Alexander Quinn and Dhariwal, Prafulla},
  booktitle={International conference on machine learning},
  pages={8162--8171},
  year={2021},
  organization={PMLR}
}

@article{bbs,
  title={A simple unpredictable pseudo-random number generator},
  author={Blum, Lenore and Blum, Manuel and Shub, Mike},
  journal={SIAM Journal on computing},
  volume={15},
  number={2},
  pages={364--383},
  year={1986},
  publisher={SIAM}
}

@article{QC,
title = {Quadratic congruential pseudorandom numbers: Distribution of lagged pairs},
journal = {Journal of Computational and Applied Mathematics},
volume = {79},
number = {1},
pages = {75-85},
year = {1997},
issn = {0377-0427},
doi = {https://doi.org/10.1016/S0377-0427(96)00158-6},
url = {https://www.sciencedirect.com/science/article/pii/S0377042796001586},
author = {Jürgen Eichenauer-Herrmann}
}

@misc{MEXP,
  title={On the Security of Modular Exponentiation with Application to the Construction of Pseudorandom Generators},
  booktitle={IACR Eprint archive},
  url={http://eprint.iacr.org/2000/064},
  note={ oded@narkis.wisdom.weizmann.ac.il 11298 received 7 Dec 2000},
  author={Oded Goldreich and Vered Rosen},
  year=2000
}

@article{MT,
author = {Matsumoto, Makoto and Nishimura, Takuji},
title = {Mersenne twister: a 623-dimensionally equidistributed uniform pseudo-random number generator},
year = {1998},
issue_date = {Jan. 1998},
publisher = {Association for Computing Machinery},
address = {New York, NY, USA},
volume = {8},
number = {1},
issn = {1049-3301},
url = {https://doi.org/10.1145/272991.272995},
doi = {10.1145/272991.272995},
journal = {ACM Trans. Model. Comput. Simul.},
month = jan,
pages = {3–30},
numpages = {28}
}

@article{LQ1,
    author = {Thomson, W. E.},
    title = {A Modified Congruence Method of Generating Pseudo-random Numbers},
    journal = {The Computer Journal},
    volume = {1},
    number = {2},
    pages = {83-83},
    year = {1958},
    month = {01},
    issn = {0010-4620},
    doi = {10.1093/comjnl/1.2.83},
    url = {https://doi.org/10.1093/comjnl/1.2.83},
    eprint = {https://academic.oup.com/comjnl/article-pdf/1/2/83/1175362/010083.pdf},
}

@book{randu,
  title={The art of computer programming},
  author={Knuth, Donald Ervin},
  volume={3},
  year={1997},
  publisher={Pearson Education}
}

@misc{bbs2,
  title={Blum Blum Shub on the GPU},
  author={Olsson, Mikael and Gullberg, Niklas},
  year={2012}
}

@article{aes1,
  title={When AES blinks: introducing optical side channel},
  author={Ferrigno, Julie and Hlav{\'a}{\v{c}}, M},
  journal={IET Information Security},
  volume={2},
  number={3},
  pages={94--98},
  year={2008},
  publisher={IET}
}

@inproceedings{aes2,
  title={A collision-attack on AES: Combining side channel-and differential-attack},
  author={Schramm, Kai and Leander, Gregor and Felke, Patrick and Paar, Christof},
  booktitle={Cryptographic Hardware and Embedded Systems-CHES 2004: 6th International Workshop Cambridge, MA, USA, August 11-13, 2004. Proceedings 6},
  pages={163--175},
  year={2004},
  organization={Springer}
}

@incollection{truerng1,
  title={True random number generators},
  author={Stip{\v{c}}evi{\'c}, Mario and Ko{\c{c}}, {\c{C}}etin Kaya},
  booktitle={Open problems in mathematics and computational science},
  pages={275--315},
  year={2014},
  publisher={Springer}
}

@article{truerng2,
  title={A survey on true random number generators based on chaos},
  author={Yu, Fei and Li, Lixiang and Tang, Qiang and Cai, Shuo and Song, Yun and Xu, Quan},
  journal={Discrete Dynamics in Nature and Society},
  volume={2019},
  number={1},
  pages={2545123},
  year={2019},
  publisher={Wiley Online Library}
}

@article{reram1,
  title={Resistive random access memory (ReRAM) based on metal oxides},
  author={Akinaga, Hiroyuki and Shima, Hisashi},
  journal={Proceedings of the IEEE},
  volume={98},
  number={12},
  pages={2237--2251},
  year={2010},
  publisher={IEEE}
}

@article{chua1,
  title={Resistance switching memories are memristors},
  author={Chua, Leon},
  journal={Handbook of memristor networks},
  pages={197--230},
  year={2019},
  publisher={Springer}
}

@article{chua2,
  title={Memristor-the missing circuit element},
  author={Chua, Leon},
  journal={IEEE Transactions on circuit theory},
  volume={18},
  number={5},
  pages={507--519},
  year={1971},
  publisher={IEEE}
}

@book{ermentrout,
  title={Mathematical foundations of neuroscience},
  author={Ermentrout, Bard and Terman, David Hillel},
  volume={35},
  year={2010},
  publisher={Springer}
}

@article{nicola1,
  title={Supervised learning in spiking neural networks with FORCE training},
  author={Nicola, Wilten and Clopath, Claudia},
  journal={Nature communications},
  volume={8},
  number={1},
  pages={2208},
  year={2017},
  publisher={Nature Publishing Group UK London}
}

@article{nicola2,
  title={A diversity of interneurons and Hebbian plasticity facilitate rapid compressible learning in the hippocampus},
  author={Nicola, Wilten and Clopath, Claudia},
  journal={Nature Neuroscience},
  volume={22},
  number={7},
  pages={1168--1181},
  year={2019},
  publisher={Nature Publishing Group US New York}
}

@article{des,
  title={The Data Encryption Standard (DES) and its strength against attacks},
  author={Coppersmith, Don},
  journal={IBM journal of research and development},
  volume={38},
  number={3},
  pages={243--250},
  year={1994},
  publisher={IBM}
}

@article{aes,
  title={Advanced encryption standard},
  author={Rijmen, Vincent and Daemen, Joan},
  journal={Proceedings of federal information processing standards publications, national institute of standards and technology},
  volume={19},
  pages={22},
  year={2001}
}

@InProceedings{slurm,
author="Jette, Morris A.
and Wickberg, Tim",
editor="Klus{\'a}{\v{c}}ek, Dalibor
and Corbal{\'a}n, Julita
and Rodrigo, Gonzalo P.",
title="Architecture of the Slurm Workload Manager",
booktitle="Job Scheduling Strategies for Parallel Processing",
year="2023",
publisher="Springer Nature Switzerland",
address="Cham",
pages="3--23",
isbn="978-3-031-43943-8"
}

@article{cycle1,
  title={Periodic orbits in chaotic systems simulated at low precision},
  author={Kl{\"o}wer, Milan and Coveney, Peter V and Paxton, E Adam and Palmer, Tim N},
  journal={Scientific Reports},
  volume={13},
  number={1},
  pages={11410},
  year={2023},
  publisher={Nature Publishing Group UK London}
}

@article{qtrng1,
  title={A 3$\mu$W CMOS True Random Number Generator With Adaptive Floating-Gate Offset Cancellation},
  author={Holleman, Jeremy and Bridges, Seth and Otis, Brian P and Diorio, Chris},
  journal={IEEE Journal of Solid-State Circuits},
  volume={43},
  number={5},
  pages={1324--1336},
  year={2008},
  publisher={IEEE}
}

@article{qtrng2,
  title={A 82-nW chaotic map true random number generator based on a sub-ranging SAR ADC},
  author={Kim, Minseo and Ha, Unsoo and Lee, Kyuho Jason and Lee, Yongsu and Yoo, Hoi-Jun},
  journal={IEEE Journal of Solid-State Circuits},
  volume={52},
  number={7},
  pages={1953--1965},
  year={2017},
  publisher={IEEE}
}

@article{chaos_map_1,
  title={n-dimensional hyperchaotic discrete map with desired positive Lyapunov exponents and application to UART secure communication},
  author={Xu, Bo and Tang, Zhongmin and Ye, Xiaoxuan and Chen, Kai and Gou, Xuan and Zhao, Jia},
  journal={Nonlinear Dynamics},
  volume={113},
  number={8},
  pages={9061--9079},
  year={2025},
  publisher={Springer}
}

@article{chaos_map_2,
  title={A 4d trigonometric-based memristor hyperchaotic map to ultra-fast prng},
  author={Xu, Bo and Luo, Xiaowei and Wang, Yifan and Bai, Libing and Chen, Kai and Zhao, Jia},
  journal={IEEE Transactions on Industrial Informatics},
  volume={20},
  number={6},
  pages={8673--8683},
  year={2024},
  publisher={IEEE}
}

@article{force1,
  title={Generating coherent patterns of activity from chaotic neural networks},
  author={Sussillo, David and Abbott, Larry F},
  journal={Neuron},
  volume={63},
  number={4},
  pages={544--557},
  year={2009},
  publisher={Elsevier}
}

@article{maass1,
  title={Real-time computing without stable states: A new framework for neural computation based on perturbations},
  author={Maass, Wolfgang and Natschl{\"a}ger, Thomas and Markram, Henry},
  journal={Neural computation},
  volume={14},
  number={11},
  pages={2531--2560},
  year={2002},
  publisher={MIT Press}
}

@article{res1,
  title={Emerging opportunities and challenges for the future of reservoir computing},
  author={Yan, Min and Huang, Can and Bienstman, Peter and Tino, Peter and Lin, Wei and Sun, Jie},
  journal={Nature Communications},
  volume={15},
  number={1},
  pages={2056},
  year={2024},
  publisher={Nature Publishing Group UK London}
}

@article{res2,
  title={A neural machine code and programming framework for the reservoir computer},
  author={Kim, Jason Z and Bassett, Dani S},
  journal={Nature Machine Intelligence},
  volume={5},
  number={6},
  pages={622--630},
  year={2023},
  publisher={Nature Publishing Group UK London}
}

@article{munoz2020two,
  title={Two new asymmetric Boolean chaos oscillators with no dependence on incommensurate time-delays and their circuit implementation},
  author={Munoz-Pacheco, Jesus M and Garc{\'\i}a-Ch{\'a}vez, Tonatiuh and Gonzalez-Diaz, Victor R and de La Fuente-Cortes, Gisela and G{\'o}mez-Pav{\'o}n, Luz del Carmen},
  journal={Symmetry},
  volume={12},
  number={4},
  pages={506},
  year={2020},
  publisher={MDPI}
}

@article{sato2021consideration,
  title={Consideration for affects of an xor in a random number generator using ring oscillators},
  author={Sato, Ryoichi and Kodera, Yuta and Ali, Md Arshad and Kusaka, Takuya and Nogami, Yasuyuki and Morelos-Zaragoza, Robert H},
  journal={Entropy},
  volume={23},
  number={9},
  pages={1168},
  year={2021},
  publisher={MDPI}
}

@article{Li2023AppliedPhysics,
    author = {Li, X. H. and Zhao, M. K. and Zhang, R. and Wan, C. H. and Wang, Y. Z. and Luo, X. M. and Liu, S. Q. and Xia, J. H. and Yu, G. Q. and Han, X. F.},
    title = {True random number generator based on spin–orbit torque magnetic tunnel junctions},
    journal = {Applied Physics Letters},
    volume = {123},
    number = {14},
    pages = {142403},
    year = {2023},
    month = {10},
    issn = {0003-6951},
    doi = {10.1063/5.0171768},
    url = {https://doi.org/10.1063/5.0171768},
    eprint = {https://pubs.aip.org/aip/apl/article-pdf/doi/10.1063/5.0171768/18150484/142403_1_5.0171768.pdf},
}

@article{Naveen2026,
doi = {10.1088/1742-6596/3191/1/012029},
url = {https://doi.org/10.1088/1742-6596/3191/1/012029},
year = {2026},
month = {apr},
publisher = {IOP Publishing},
volume = {3191},
number = {1},
pages = {012029},
author = {Naveen Srimal M and Kishore S and Banu, Aashiq and Muthaiah R and Sivaraman R},
title = {From Chaos to Cryptography: A Memristive True Random Number Generator with NIST-Compliant Output},
journal = {Journal of Physics: Conference Series}
}

@ARTICLE{Yu2021Design,
AUTHOR={Yu, Fei  and Zhang, Zinan  and Shen, Hui  and Huang, Yuanyuan  and Cai, Shuo  and Jin, Jie  and Du, Sichun },
TITLE={Design and FPGA Implementation of a Pseudo-random Number Generator Based on a Hopfield Neural Network Under Electromagnetic Radiation},
JOURNAL={Frontiers in Physics},
VOLUME={Volume 9 - 2021},
YEAR={2021},
URL={https://www.frontiersin.org/journals/physics/articles/10.3389/fphy.2021.690651},
DOI={10.3389/fphy.2021.690651},
ISSN={2296-424X}}

@article{jia2026widening,
  title={Widening the High-Entropy Operating Window of Perpendicular Magnetic Tunnel Junction TRNGs via Multibit xor Whitening},
  author={Jia, Qi and Egan, Shannon and Lv, Yang and Wang, Jian-Ping},
  journal={IEEE Transactions on Electron Devices},
  year={2026},
  publisher={IEEE}
}

@article{vodenicarevic2017low,
  title={Low-energy truly random number generation with superparamagnetic tunnel junctions for unconventional computing},
  author={Vodenicarevic, Damir and Locatelli, Nicolas and Mizrahi, Alice and Friedman, Joseph S and Vincent, Adrien F and Romera, Miguel and Fukushima, Akio and Yakushiji, Kay and Kubota, Hitoshi and Yuasa, Shinji and others},
  journal={Physical Review Applied},
  volume={8},
  number={5},
  pages={054045},
  year={2017},
  publisher={APS}
}

\end{document}